\documentclass[10pt,journal,compsoc]{IEEEtran}
\usepackage{times}
\usepackage{epsfig}
\usepackage{graphicx}
\usepackage{amsmath}
\usepackage{amssymb}
\usepackage{soul}
\usepackage{color}
\usepackage{booktabs}
\usepackage{enumitem}
\usepackage{multirow}
\usepackage{array}
\usepackage{cite}
\usepackage{amsmath}
\usepackage{algorithm}
\usepackage{algorithmic}
\usepackage{threeparttable}
\usepackage[section]{placeins}
\usepackage{amsfonts}       
\usepackage{nicefrac}       
\usepackage{microtype}      
\usepackage{xcolor}         
\usepackage{diagbox}
\usepackage{pifont}
\usepackage{multicol}
\usepackage{enumerate}

\usepackage[pagebackref=true,breaklinks=true,colorlinks,bookmarks=false]{hyperref}

\makeatletter
\def\hlinew#1{%
 \noalign{\ifnum0=`}\fi\hrule \@height #1 \futurelet
 \reserved@a\@xhline}
\makeatother%

\newcommand{\PreserveBackslash}[1]{\let\temp=\\#1\let\\=\temp}
\newcolumntype{C}[1]{>{\PreserveBackslash\centering}p{#1}}
\newcolumntype{L}[1]{>{\PreserveBackslash\raggedleft}p{#1}}
\newcolumntype{R}[1]{>{\PreserveBackslash\raggedright}p{#1}}

\begin{document}

\title{Unsupervised Deraining: Where Asymmetric Contrastive Learning Meets Self-similarity}

\author{Yi~Chang,~\IEEEmembership{Member,~IEEE,}
        Yun~Guo,
        Yuntong~Ye,
        Changfeng~Yu,
        Lin~Zhu,
        Xile~Zhao,
        Luxin~Yan$^*$,~\IEEEmembership{Member,~IEEE},
        and~Yonghong~Tian,~\IEEEmembership{Fellow,~IEEE}
}

\IEEEtitleabstractindextext{%
\begin{abstract}
  Most of the existing learning-based deraining methods are supervisedly trained on synthetic rainy-clean pairs. The domain gap between the synthetic and real rain makes them less generalized to complex real rainy scenes.
  Moreover, the existing methods mainly utilize the property of the image or rain layers independently, while few of them have considered their mutually exclusive relationship. To solve above dilemma, we explore the intrinsic intra-similarity within each layer and inter-exclusiveness between two layers and propose an unsupervised non-local contrastive learning (NLCL) deraining method. The non-local self-similarity image patches as the positives are tightly pulled together and rain patches as the negatives are remarkably pushed away, and vice versa. On one hand, the intrinsic self-similarity knowledge within positive/negative samples of each layer benefits us to discover more compact representation; on the other hand, the mutually exclusive property between the two layers enriches the discriminative decomposition. Thus, the internal self-similarity within each layer (\emph{similarity}) and the external exclusive relationship of the two layers (\emph{dissimilarity}) serving as a generic image prior jointly facilitate us to unsupervisedly differentiate the rain from clean image. We further discover that the intrinsic dimension of the non-local image patches is generally higher than that of the rain patches. This motivates us to design an asymmetric contrastive loss to precisely model the compactness discrepancy of the two layers for better discriminative decomposition. In addition, considering that the existing real rain datasets are of low quality, either small scale or downloaded from the internet, we collect a large-scale real dataset under various rainy weathers that contains high-resolution rainy images. Extensive experiments on different real rainy datasets demonstrate that the proposed method obtains state-of-the-art performance in real deraining. Both the code and the newly collected datasets will be available at https://owuchangyuo.github.io.
\end{abstract}

}


\maketitle

\IEEEdisplaynontitleabstractindextext
\IEEEpeerreviewmaketitle

\IEEEraisesectionheading{\section{Introduction}\label{sec:introduction}}
\IEEEPARstart{T}{he} existing high-level computer vision tasks such as image segmentation \cite{chen2018encoder}, and object detection \cite{liu2016ssd} have achieved significant progress in recent years. Unfortunately, their performance would suffer from degradation under the rainy weather \cite{bahnsen2018rain, jiang2020multi, li2019rainflow}. To alleviate the influence of the rain, numerous full-supervised deraining methods have been proposed \cite{fu2017clearing, yang2017deep, zhang2018density}. Although they can achieve good results on simulated rainy image, they cannot well generalize to the real rain because of the domain gap between the simplified synthetic rain and complex real rain \cite{ye2021closing}. The goal of this work is to remove the real rain in an unsupervised manner.

To handle the real-world complex rainy images, the optimization-based methods are firstly proposed with hand-crafted priors such as sparse coding \cite{luo2015removing}, low-rank \cite{chang2017transformed} and Gaussian mixture model \cite{li2016rain}. However, these hand-crafted priors are of limited representation ability, especially for highly complex and varied rainy scenes. To rectify this weakness, the learning-based CNN methods \cite{fu2017clearing, yang2017deep, li2018recurrent, li2019heavy} have made great progresses. The researchers starting from the supervised learning methods try their best to simulate the rain as real as possible with sophisticated models, such as additive model \cite{kang2011automatic}, screen blend model \cite{luo2015removing}, heavy rain model \cite{yang2017deep}, comprehensive rain model \cite{hu2019depth}, rendering model \cite{halder2019physics}, and learned rain models \cite{wang2021rain, ni2021controlling, ye2021closing} to name a few.

\begin{figure*}[htbp]
  \centering
		\includegraphics[width=1.0\textwidth]{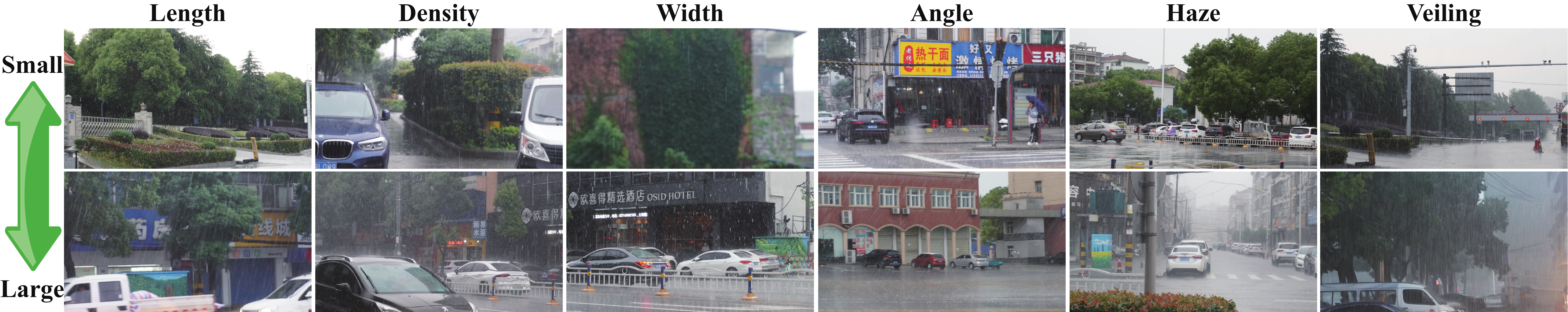}
		\caption{Visualization of various kind of rainy images in our real dataset. This high-quality dataset covers various rain categories with different attributes.}
  \label{DataSetVisualization}
\end{figure*}

However, real rain is related with various factors which is impossible to be comprehensively considered. The appearance of the rain is closely associated with the camera exposure time (length), rainfall amount (density), raindrop size (width), wind direction (angle), and distance (haze/veiling). In Fig. \ref{DataSetVisualization}, we show the representative rains from our collected real rainy images. The real rain contains not only the rain streaks which are easier to be simulated, but also the complex veiling and haze artifacts. The veiling and haze are highly correlated with the scene semantics such as the depth, which makes it difficult to be accurately simulated.

\begin{figure*}[htbp]
  \centering
		\includegraphics[width=1.0\textwidth]{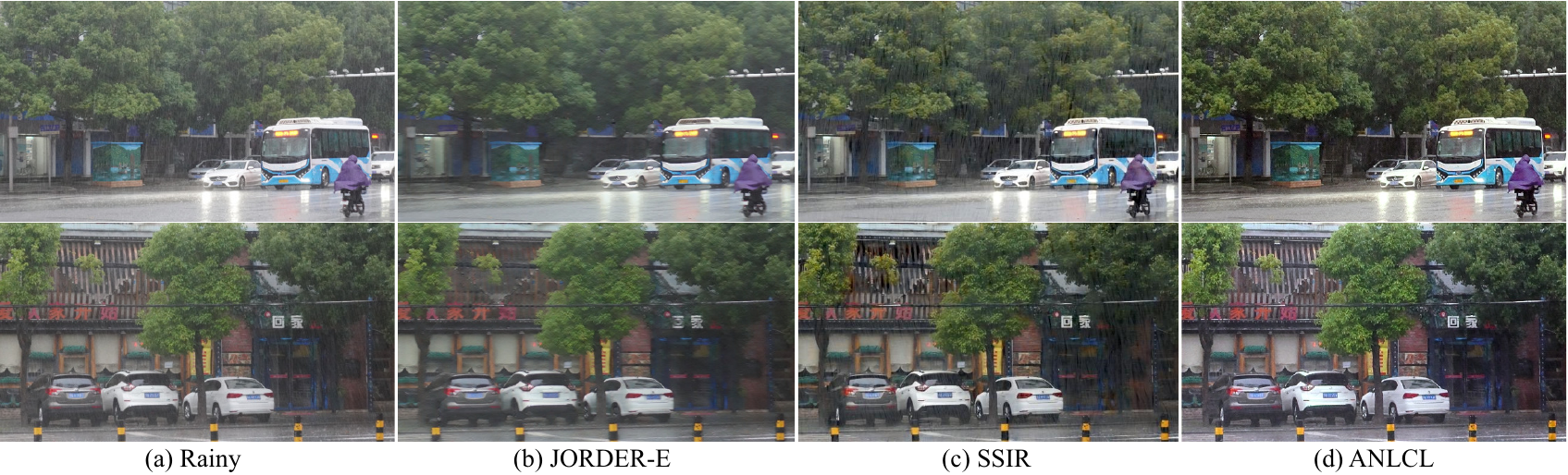}
		\caption{The influence of the domain gap between the synthetic and real rain. (a) Real rain images. (b) The supervised JORDER-E \cite{yang2020joint} usually over-smoothes the image details. (c) The semi-supervised SSIR \cite{wei2019semi} contains obvious rain residuals. (d) The proposed unsupervised ANLCL can remove not only the rain streaks but also real-world veiling/haze, meanwhile well preserve image structures. Please zoom in for better visualization. }
  \label{SupervisedFailed}
\end{figure*}

Consequently, there inevitably exist domain gap between these synthetic rain models and real rain degradation, which makes the supervised methods less generalize well to the real rain. To illustrate this issue, we provide the deraining results of representative supervised method JORDER-E \cite{yang2020joint} on real rain images in Fig. \ref{SupervisedFailed}. Although JORDER-E has achieved very impressive results on synthetic rain, it is less effective for the complex real rains with diverse appearances. The oversmooth phenomenon can be observed clearly in Fig. \ref{SupervisedFailed}(b), mainly due to the gap between the real test and synthetic heavy rain.

Latter, the semi-supervised deraining methods have been proposed to effectively improve the robustness for real rain \cite{wei2019semi, yasarla2020syn2real, wei2021deraincyclegan, ye2021closing, huang2021memory, liu2021unpaired}, where they employ the simulated labels for good initialization and unlabeled real rain for generalization. Their performances still depend on the distribution gap between the simulated and real rainy images to some extent. Once the distributions are of large distance, the semi-supervised deraining result by SSIR \cite{wei2019semi} would be less satisfactory, as shown in Fig. \ref{SupervisedFailed}(c). The unsupervised methods have raised more attentions for real rain removal, mainly including the CycleGAN-based unpaired image translation methods \cite{zhu2019singe, jin2019unsupervised, wei2021deraincyclegan, chen2022unpaired} and the optimization-model driven deep prior network \cite{yu2021unsupervised}. In Fig. \ref{RelatedWork}, we summarize the development of the single image deraining methods. The previous methods including the unsupervised ones mainly pay attention to the property of the image or rain layer independently, yet seldom consider the mutually exclusive relationship between the two layers.

To overcome these problems, we formulate the image deraining into a novel non-local contrastive decomposition framework, which aims at decomposing the rainy image into two distinguishable layers: clean image layer and rain layer, as shown in Fig. \ref{Motivation}. On one hand, we not only take advantage of the non-local self-similarity properties within both image and rain layers, benefiting us to learn compact representation for each layer; on the other hand, we model the mutually exclusive relationship between the two layers so as to enrich the discriminative representation. Thus, the internal self-similarity within each layer and the external exclusive relationship of the two layers allow us free from the supervision, and jointly facilitate us to differentiate the rain from clean image. Note that, we equally treat the image and rain layers as both positive and negative, and propose the bidirectionally symmetric contrastive learning for better decomposition.

To the best of our knowledge, we are the first to incorporate non-local self-similarity into contrastive learning for positive/negative sampling. The advantage of the proposed non-local sampling is twofold. First, the non-local self-similarity sampling strategy would naturally guarantee more compact clusters for positives and negatives respectively, which would benefit us to differ the positives from negatives. Second, these positive non-local patches are  the samples searched from real images with diverse variable information, not manually generated fake samples, which would provide more faithful information for representation. Moreover, compared with the image-level samples, the patches would greatly enrich the sample numbers for better contrastive learning. Note that, the non-local strategy is not only applicable for the positive samples, but also beneficial to the negative samples. In addition, we provide an guidance of how to design a good encoder for better embedding in contrastive learning.

This work is an extension of our earlier publication in CVPR 2022 \cite{ye2022unsupervised}. The main extensions are three folds. In this work, we further analyze the intrinsic dimension discrepancy between the image and rain non-local clusters. Different from the previous version where we equally treat the image and rain samples in the decomposition, in this work we design an asymmetric contrastive loss to precisely model the compactness discrepancy of the two layers for better discriminative decomposition. Second, we construct a large-scale high-quality real rainy image dataset with diverse rain appearances through field collection, considering that most of the existing real rain datasets are collected from the internet with poor quality or with limited scenes. Third, more qualitative and quantitative experiments including the advantage of non-local sampling and promotion for downstream detection are conducted. We demonstrate that the ANLCL is a general prior and can be generalized to other bad weathers tasks and also embedded into previous methods with sufficient improvement. Overall, our contributions can be summarized as follow:

 \begin{figure*}[htbp]
  \centering
     \includegraphics[width=1.00\linewidth]{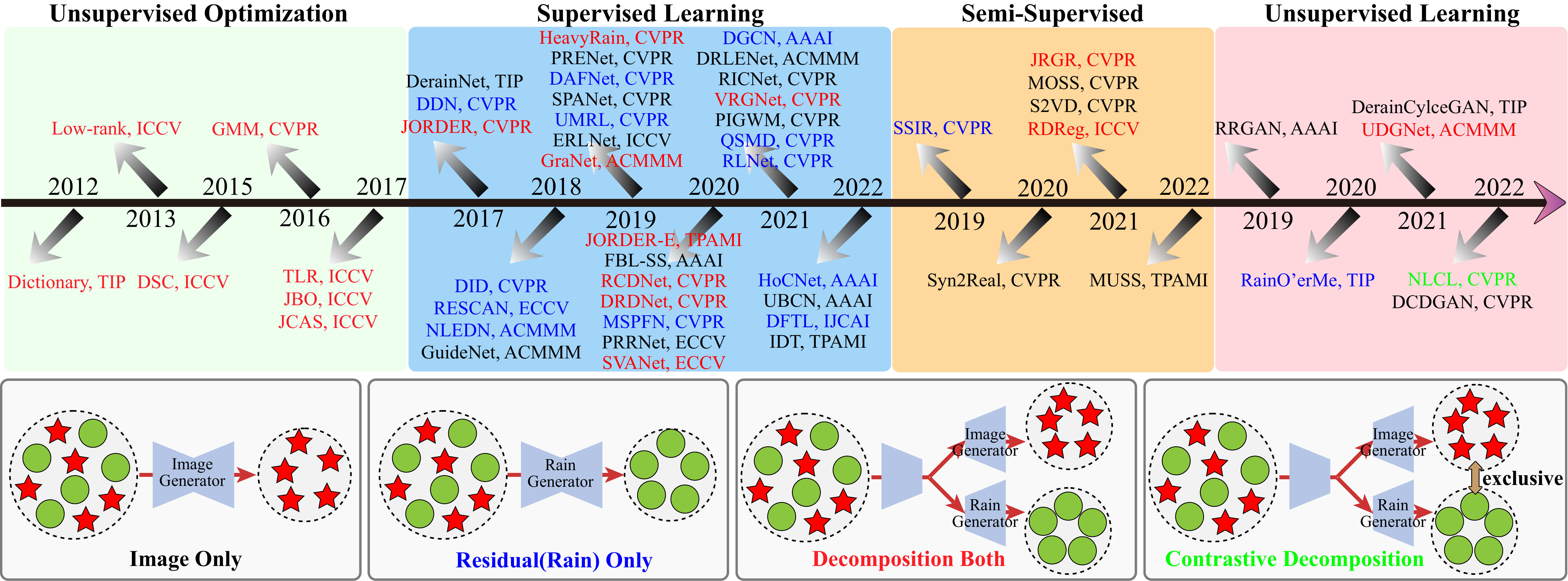}
  \caption{The development of the single image deraining methods. The time period up to 2017 is dominated by model-driven optimization methods, with methods after 2017 dominated by data-driven learning methods: supervised methods, semi-supervised methods and unsupervised methods.}
  \label{RelatedWork}
\end{figure*}

\begin{itemize}[leftmargin=10pt]
\item We formulate the single image rain removal into an unsupervised contrastive decomposition framework, and propose a novel non-local contrastive learning (NLCL) deraining method which simultaneously explore the intrinsic intra-similarity within each layer and inter-exclusiveness between two layers. Our work is the first to explicitly consider the exclusive relationship between the rain and image in the learning network.
\item We connect the contrastive learning with the non-local self-similarity. Instead of the conventional instance/image-level sampling, we demonstrate that the non-local patch-level sampling strategy naturally endows the positive/negative samples with more compact and discriminative representation for better decomposition. In addition, we provide an guidance of how to design a good encoder for better embedding.
 \item We discover the asymmetric property within the image and rain spaces: the intrinsic dimension of the non-local image patches is generally higher than that of the rain patches. We extend the symmetric NLCL to asymmetric one by designing an asymmetric contrastive loss to capture this discrepancy. We show how this asymmetric property would benefit us to improve the discriminative representation for better decomposition.
 \item We release a large-scale high-quality real rain image dataset. Our images are collected in different rain weathers on the city roads with diverse rain appearances and abundant traffic elements annotation. This real dataset would be a good testbed for the community, especially for the unsupervised deraining methods. We conduct extensive experiments on both synthetic and real-world rain datasets, and show that ANLCL outperforms favorably state-of-the-art methods on real image deraining.
\end{itemize}

\section{Related Work}
\noindent
\textbf{Single Image Deraining.} In Fig. \ref{RelatedWork}, we provide a brief development of the single image deraining including the supervised, semi-supervised and unsupervised methods. The interested reader could refer to survey work \cite{Yang2020Single, wang2022survey} for detailed description.

Most of the existing methods are full-supervised which require a large number of paired rainy and clean images as training samples \cite{yang2017deep, fu2017removing, zhu2017joint, zhang2018density, li2018recurrent, li2018non, fan2018residual, hu2019depth, li2019heavy, ren2019progressive, wang2019erl, yasarla2019uncertainty, yang2020towards, deng2020detail, jiang2020multi, zhang2020beyond, wang2020rethinking, wang2021multi, fu2021rain, zhou2021image, chen2021robust, yi2021structure, quan2021removing, fu2021rain, wang2022online, wu2022decoder, li2022close}. The seminal learning-based CNN works for image deraining is proposed by Fu \emph{el at}. \cite{fu2017removing} and Yang \emph{et al}. \cite{yang2017deep}. Fu \emph{et al}. \cite{fu2017removing} introduced the end-to-end CNN with residual learning for rain streaks removal. JORDER-E \cite{yang2017deep} jointly learned the rain detection and removal in a multi-task network with progressive guidance. Latter, the multi-stage \cite{yang2017deep}, multi-scale \cite{jiang2020multi}, density \cite{zhang2018density}, and attention \cite{li2018recurrent} have been widely utilized for better representation. The interested readers could refer to survey \cite{Yang2020Single} for comprehensive description. Among them, the most typical CNN and Transformer based methods are JORDER-E \cite{yang2020joint} and IDT \cite{xiao2022image}, respectively. IDT \cite{xiao2022image} is a very recently proposed deraining method to capture the long-range dependencies of the rains. Those works have made great progress for the community. Unfortunately, the domain gap between the complex real rain and the simplified synthetic rain would limit their generalization in real scenes.

The semi-supervised deraining models \cite{wei2019semi, yasarla2020syn2real, wei2021deraincyclegan, ye2021closing, yue2021semi, huang2021memory, huang2022memory} are proposed by additionally introducing the real dataset for better generalization. For example, Wei \emph{et al}. \cite{wei2019semi} proposed the first semi-supervised transfer learning framework via network structure/weight sharing to better utilize unlabeled real images. Latter, Yasarla \emph{et al}. \cite{yasarla2020syn2real} presented the Gaussian process-based semi-supervised learning for real image deraining. Recently, Huang \emph{et al}. \cite{huang2022memory} developed memory-uncertainty guided semi-supervised (MUSS) learning framework that is equipped with memory modules to generate the prototypical (pseudo labels) rain patterns. Although these semi-supervised methods could alleviate this issue to some extent, the choice of the synthetic datasets would heavily determine the final real performance.

Recently, the unsupervised deraining methods have emerged \cite{zhu2019singe, jin2019unsupervised, wei2021deraincyclegan, yu2021unsupervised, chen2022unpaired, ye2022unsupervised}. Most of the previous unsupervised works formulated the unsupervised image deraining as the image generation task via the generative adversarial learning. Wei \emph{et al}. \cite{wei2021deraincyclegan} extended the classical CycleGAN into the DerainCycleGAN using unpaired data for real image deraining. Yu \emph{et al}. \cite{yu2021unsupervised} took the prior knowledge of the rain streak into consideration, and connected the model-driven and data-driven methods via an unsupervised learning framework. In this work, we propose a novel contrastive learning framework for unsupervised deraining. Compared with previous methods, the ANLCL could further take mutual exclusive relationship between image and rain layers into consideration.

\noindent
\textbf{Contrastive Learning.} Contrastive learning (CL) has achieved promising results in unsupervised representation learning \cite{chen2020simple, he2020momentum, chen2020improved, oord2018representation, wu2018unsupervised, chen2020big}. The main idea is to push the features of unrelated data (as negatives) and pull the related data (as positives), so as to learn the representations which are discriminative to the negatives and invariant between the positives. CL can be effectively applied by appropriately defining the positives and negatives in terms of the tasks, including the multi-views \cite{tian2019contrastive, 2020What}, temporal coherence in video sequence \cite{Han2020VideoRepresentation}, augmented transformation \cite{chen2020simple, he2020momentum}, to name a few. Recently, researches have applied the CL to low-level applications \cite{park2020contrastive, liu2021divco, wu2021contrastive, xia2022efficient, liang2022semantically, chen2022unpaired, liu2022twin, chen2022learning, wang2022ucl}. Most of the existing CL methods take the clean images as the positive and the degraded images as the negative samples. For example, Wu \emph{et al}. \cite{wu2021contrastive} pulled the restored image closer to ground truth (GT) and pushed them far away from the hazy image in the representation space within a supervised framework. Latter, following the similar sampling strategy, the authors \cite{chen2022unpaired, liu2022twin, chen2022learning, wang2022ucl} extend it to the unsupervised framework by combined the adversarial learning with contrastive learning.

Our ANLCL is significantly different from \cite{wu2021contrastive, chen2022unpaired, liu2022twin, chen2022learning, wang2022ucl} in two aspects. First, the key positive/negative sampling strategy is different. The proposed ANLCL take the image layer and degraded rain layer as the positive and negative, respectively. Compared with previous clean and degraded image samplings \cite{wu2021contrastive, chen2022unpaired, liu2022twin, chen2022learning, wang2022ucl}, the image layer is entangled with the degraded layer which makes it ambiguous to learn the discriminative features to differ them from each other. On the contrary, the proposed ANLCL employs explicit disentanglement between the two layers, in which the two layers are with distinct discrepancy patterns. The rain layer is relative simpler with repetitive line-patterns and the image is with meaningfully geometrical structures. Intuitively, the disentanglement between the rain and image layer would better facilitate us to achieve the final goal: decouple the rain from the clean image. Second, the previous CL methods \cite{wu2021contrastive, chen2022unpaired, liu2022twin, chen2022learning, wang2022ucl} consistently employ the instance image-level samples for contrast, while we have explored the intrinsic similarity between the patches within a single image. Compared with the image samples, the intrinsic self-similarity within the positive or negative samples would significantly ease the learning procedure and boost more compact feature space. Moreover, non-local patches would greatly enrich the sample numbers for discriminative feature learning.

\noindent
\textbf{Image Decomposition for Deraining.} According to the estimated output, we can classify the existing deraining methods into three categories: image-based, rain (residual)-based, and decomposition-based. In 2012 to 2017, the image deraining field is dominated by the image decomposition based model-driven optimization methods \cite{kang2012automatic, Chen2013A, luo2015removing, li2016rain, zhu2017joint, chang2017transformed, gu2017joint}. For example, the pioneer work \cite{kang2012automatic} introduced the dual sparse dictionaries for both rain and non-rain component representation. The main idea of image decomposition based deraining is to make the two components lie on two different subspaces, so as to decouple the rainy image into the image and rain layer. The advantage of the decomposition-based method over the single rain/image-based method is that more domain knowledge can be utilized for better discriminative feature extraction. Moreover, the relationship between the two components can be further modeled to benefit from each other.

With the advent of the end-to-end CNN, it is intuitive to employ the CNN directly mapping the rainy image to clean image \cite{fu2017clearing, ren2019progressive, xiao2022image}, due to its simplicity and powerful representation. Latter, considering the rain can be regarded as the sparse residual error, the researchers have proposed the rain-based residual learning methods \cite{fu2017removing, li2018recurrent, deng2020detail} which significantly reduce the training difficulty. However, both the rain-based and image-based method have ignored the relationship between the two components. The decomposition-based CNN methods have achieved state-of-the-art performance such as the well-known JORDER \cite{yang2017deep, yang2020joint} and RCDNet \cite{Wang2020Model}. In \cite{yang2020joint}, the rain layer is estimated as the location guidance for the image layer estimation, which serve as a rain and non-rain region attention for better image estimation. Wang \emph{et al}. \cite{Wang2020Model} proposed an interpretable network architecture by unfolding the decomposition model into a rain convolutional dictionary network (RCDNet).

In this work, we advance the image decomposition paradigm from two aspects. First, we bridge contrastive learning with the decomposition framework. Compared with previous work, the proposed method not only exploits both the image and rain properties, but also take the mutual exclusion relationship between the two layers into consideration. The contrastive learning could naturally model both the intra-similarity within each layer (\emph{self-similarity}) and inter-exclusiveness between two layers (\emph{dissimilarity}). Second, most of the existing learning-based decomposition deraining methods require the synthetic clean and degraded pair in a supervised manner. On the contrary, the proposed method explores the \emph{self-similarity} and \emph{dissimilarity} as a generic prior via unsupervised learning, which ensures the proposed method generalize well for the real rain.

\noindent
\textbf{Non-local Self-similarity.} The nonlocal prior reveals a general image property that the similar small patches tend to recurrently appeared within a single image. This generic property could provide group sparsity of the image with structural representation. The self-similarity serves as a powerful image prior model, which has been demonstrated in various image restoration techniques including filtering methods \cite{buades2005non, dabov2007image}, sparse optimization models \cite{mairal2009non, gu2014weighted}, and deep neural networks \cite{liu2018non, wang2018non, bell2019blind}. Beneficial from capturing the correlation among the self-similarity patches, these non-local based methods have achieved the state-of-the-art performances at that time, such as the BM3D in denoising \cite{dabov2007image}, WNNM in restoration \cite{gu2014weighted}, and kernelGAN in blind super-resolution \cite{bell2019blind}. The self-similarity is a very generic prior for unsupervised learning. For example, Krull \emph{et al}. \cite{krull2019noise2void} learned to predict the center pixel according to its local neighborhood, which made use of the redundant property of the images while the random noise can be removed during the self-regression procedure. In this work, we further show that the non-local self-similarity can be served as powerful prior for unsupervised learning, and how it benefits the contrastive learning in terms of the positive/negative sampling, and boosts the performance in low-level image deraining task.

\section{Field Collection Real Rain Dataset}
The datasets play an important role in deep learning era. The researchers have made great progress to provide numerous rain datasets, mainly including the synthetic and real rains. We summary the existing typical rain datasets in Table \ref{DatasetComparison}. Most of the existing datasets are synthetic-based. The pioneer work was proposed by Nayar and Garg \cite{garg2005does, garg2007vision} with geometric and photometric analysis for real rain appearance. According to the photometric model, Fu \emph{et al}. \cite{fu2017removing} mimicked the rain imaging procedure via Photoshop. Further, Yang \emph{et al}. \cite{yang2017deep} and Hu \emph{et al}. \cite{hu2019depth} took the out-of-focus veiling, distant haze effect, and close rain occlusion into a comprehensive rain model. These synthetic models have greatly promoted the development of this field. Although these complicated synthetic models can simulate the rain effect to some extent, there still suffers from the domain shift issue between the synthetic and real rain. To handle the real complex rain, there are several real rain datasets in recent years. We describe the proposed real rain dataset and compare with previous datasets in three aspects.

\begin{table*}[htbp]
\centering
\tiny
\caption{Summary of existing synthetic and real rain datasets.}
\renewcommand\arraystretch{1.15}
\begin{tabular}{C{0.6cm}|C{1.39cm}|C{0.8cm}|C{0.85cm}|C{0.80cm}|C{0.8cm}|C{0.65cm}|C{0.9cm}|C{0.75cm}|C{2.90cm}|C{3.20cm}}
\hline
Category                   & Datasets      & Publish   & Collection                                                       & Format      & Resolution & Number  & Orientation                                                     & Annotation                                             & Highlight                                                                                                       & Limitation                                                                                                             \\ \hline
\multirow{3}{*}{Synthetic} & Rain100  \cite{yang2017deep}     & 2017CVPR  & Synthetic                                                        & Image       & 481*321    & 2000    & None                                                         & No                                                     & Multiple rain layers accumulation                                                                               & Unrealistic rain                                                                                                       \\ \cline{2-11}
                           & Rain14000  \cite{fu2017removing}   & 2017CVPR  & Synthetic                                                        & Image       & 512*384    & 14000   & None                                                         & No                                                     & \begin{tabular}[c]{@{}c@{}}Clear rain streaks generated from\\  Photoshop with diverse directions\end{tabular}  & Domain gap between real and synthetic                                                                                  \\ \cline{2-11}
                           & RainCityscape \cite{hu2019depth} & 2019CVPR  & Synthetic                                                        & Image       & 2048*1024  & 10620   & Driving                                                         & \begin{tabular}[c]{@{}c@{}}Bounding\\ Box\end{tabular} & \begin{tabular}[c]{@{}c@{}}High-quality and resolution image;\\ Abundant object with annotation\end{tabular}                      & Domain gap between real and synthetic                                                                                  \\ \hline
\multirow{7}{*}{Real}      & RIS  \cite{li2019single}         & 2019CVPR  & Internet                                                         & Video       & 640*368    & 154     & Surveillance                                                    & \begin{tabular}[c]{@{}c@{}}Bounding\\ Box\end{tabular} & \begin{tabular}[c]{@{}c@{}}Rain with mist;\\Abundant objects with annotation\end{tabular}          & \begin{tabular}[c]{@{}c@{}}Low-resolution; Rare rain streaks\\ Compression artifacts; Limited scenes\end{tabular}       \\ \cline{2-11}
                           & SPA-Data  \cite{wang2019spatial}    & 2019CVPR  & \begin{tabular}[c]{@{}c@{}}iPhone/\\ YouTube\end{tabular}        & Video       & 512*512    & 170     & Surveillance                                                    & No                                                     & \begin{tabular}[c]{@{}c@{}}Paired real rainy and clean image\\  generated from aligned video\end{tabular}       & \begin{tabular}[c]{@{}c@{}}Less heavy rain; Limited backgrounds\\ (buildings) and foreground objects; Watermarks\end{tabular}      \\ \cline{2-11}
                           & NR-IQA  \cite{wu2020subjective}      & 2020TCSVT & Internet                                                         & Image       & 1000*680   & 206     & None                                                         & No                                                     & Lossless compression format                                                                                     & Limited images; Uneven image resolution                                                                   \\ \cline{2-11}
                           & Real3000 \cite{yue2021semi}     & 2021ICCV  & Internet & Image       & 942*654    & 3000    & None                                                         & No                                                     & Rain with different appearances                                                                                 & \begin{tabular}[c]{@{}c@{}}Numerous watermarks;  Uneven image resolution;\\ Mixed up with synthetic rain (Cartoon)\end{tabular}   \\ \cline{2-11}
                           & PairedRain  \cite{ba2022not}  & 2022ECCV  & YouTube                                                          & Video       & 666*339    & 101     & Surveillance                                                    & No                                                     & \begin{tabular}[c]{@{}c@{}}Paired real rainy and clean image\\ captured from different times\end{tabular}        & \begin{tabular}[c]{@{}c@{}}Heavy compression artifacts; Mismatch\\  between paired images; Limited scenes\end{tabular} \\ \cline{2-11}
                           & SSID \cite{huang2022memory}         & 2022TPAMI & Internet                                                         & Video/Image & 1280*720    &180/950 & None                                                         & No                                                     & Large diversity of rain scenes                                                                                  & Compression artifacts; Watermarks                                                                                                \\ \cline{2-11}
                           &  FCRealRain     & --        & \begin{tabular}[c]{@{}c@{}}Sony\\ILCE-6400 \end{tabular}                          & Image       & 4240*2400  & 4000    &  Driving & \begin{tabular}[c]{@{}c@{}}Bounding\\ Box\end{tabular} & \begin{tabular}[c]{@{}c@{}}Very clear rain with diverse appearances; \\ High-quality image without compression;\\Abundant object with annotation\end{tabular} & \begin{tabular}[c]{@{}c@{}}Without clean and degraded pairs \\ for unsupervised training only\end{tabular}             \\ \hline
\end{tabular}
\label{DatasetComparison}
\end{table*}

\noindent
\textbf{Collection and Resolution}: Most of the existing real rain datasets are collected from the internet. The internet collected rainy images come from a variety of different sources including television/films, cartoon/artistic, surveillance video and so on. The resolution of these images are vastly different, ranging from 6000*3500 to 250*180. Most of the internet collected rainy images are with relative small size. We report the average resolution of each dataset in Table \ref{DatasetComparison}. Note that, SPA-Data does not report the size of the original video, where we list the cropped size 512*512 instead. We can observe that resolution of the real datasets are mostly less than that of 720p. In this work we collect the real rain images under rainy weathers with the Sony ILCE-6400 Camera. We empirically set the shutter speed between [1/160, 1/60], aperture as f/5.6, and focal length as 50mm. The spatial resolution is consistent 4240*2400, slightly larger than the standard 4K (3840*2160), which offers more details for both the background and rain.

\noindent
\textbf{Format and Number}: The real rain dataset collection is difficult, since it heavily depends on the precipitation with high degree of randomness. Moreover, the rain would cause inconvenient for the collection, due to possible damage to the electron device. That is the main reason why the number of the rainy images in previous dataset is relative small. It is worth noting that in this work we consistently report the source number of each dataset for fair comparison. The augmentation to enlarge the number such as the spatial patch cropping or temporal frame extraction is not reported. In Table \ref{DatasetComparison}, we can observe that the image/video number in synthetic rain dataset is significantly larger than that of the real rain dataset, because of the low cost of the synthetic dataset. There are two common formats to construct the real rain dataset: image and video. The number of the videos clip is usually small, such as 154 in RIS \cite{li2019single}, 170 in SPA-Data \cite{wang2019spatial}, 101 in PairedRain \cite{ba2022not}, and 180 in SSID \cite{huang2022memory}. Nevertheless, the video clips can be further extracted as the image frame, although there is temporal redundancy between each frame. In this work, we collect a relative large dataset with 4000 high-resolution source real rain images, which contains diverse scenes and different rain patterns.

\noindent
\textbf{Highlight and Limitation}: The synthetic rain datasets could controllably generate diverse rains on high-quality images with large number samples. Moreover, the synthetic datasets could easily provide the paired clean and degraded images. These large-scale paired synthetic datasets can be well utilized by supervised learning with powerful representation. However, real rain is much more complex with the domain shift between them, which makes the synthetic trained model less robust to the real rain. Thus, the real rain datasets have been proposed to advance the complex real rain removal in the real-world. The SPA-Data \cite{wang2019spatial} and PairedRain \cite{ba2022not} try to construct real rainy and clean paired images from the videos, which offers a new path to the single image real rain removal.

Existing real datasets are mostly downloaded from the internet such as Youtube. The main problem of these real rainy dataset is that they are collected from the internet which have been unexpectedly compressed during the streaming \cite{li2019single, ba2022not, huang2022memory}. The compressed videos would, on one hand result in the blocking artifacts, and on the other hand weaken the rain features. The second problem is the unexpected watermarks in the internet videos. These numerous watermarks \cite{wang2019spatial, ba2022not} may lead to a learning bias during the training procedure. What is worse, some real datasets are not well cleaned with the fake images such as cartoon or portrait. The third problem is the limited number and scene in the dataset. Most of the real rain datasets are less than 1000 source image/videos which can not match the powerful representation ability of the network. Last but not least, most of the real datasets are designed mainly for low-level image deraining, but not for downstream tasks as such detection, except for the RIS \cite{li2019single}. These datasets do not face specific application such as the typical driving or surveillance, which restricts their further applications.

In this work, the proposed dataset is Field Collection Real Rain (FCRealRain) with high-quality images and diverse rain appearance. Moreover, the FCRealRain focuses on the driving scenes, which contains abundant object on the street. We have provided the bounding box annotation for six typical categories: people, car, bus, motorcycle, traffic light and traffic sign. Thus, the downstream detection can be further employed to validate the effectiveness of the image deraining. It is worth noting that most of existing real datasets only contain the real rainy images without the corresponding clean ground truth, including the proposed FCRealRain.
\section{Asymmetric Non-local Contrastive Decomposition for Image Deraining}
\subsection{Contrastive Image Decomposition Framework}
Given a rainy image $\textbf{\emph{O}}$, our goal is to decompose the rainy image into a clean background layer ${\textbf{\emph{B}}}$ and a rain layer ${\textbf{\emph{R}}}$. The degradation procedure can be formulated as:
\begin{equation}
\setlength{\abovedisplayskip}{2pt}
\setlength{\belowdisplayskip}{2pt}
\textbf{\emph{O}} = \textbf{\emph{B}} + \textbf{\emph{R}}.
\label{eq:DegradationModel}
\end{equation}

\begin{figure*}[htbp]
  \centering
     \includegraphics[width=1.0\linewidth]{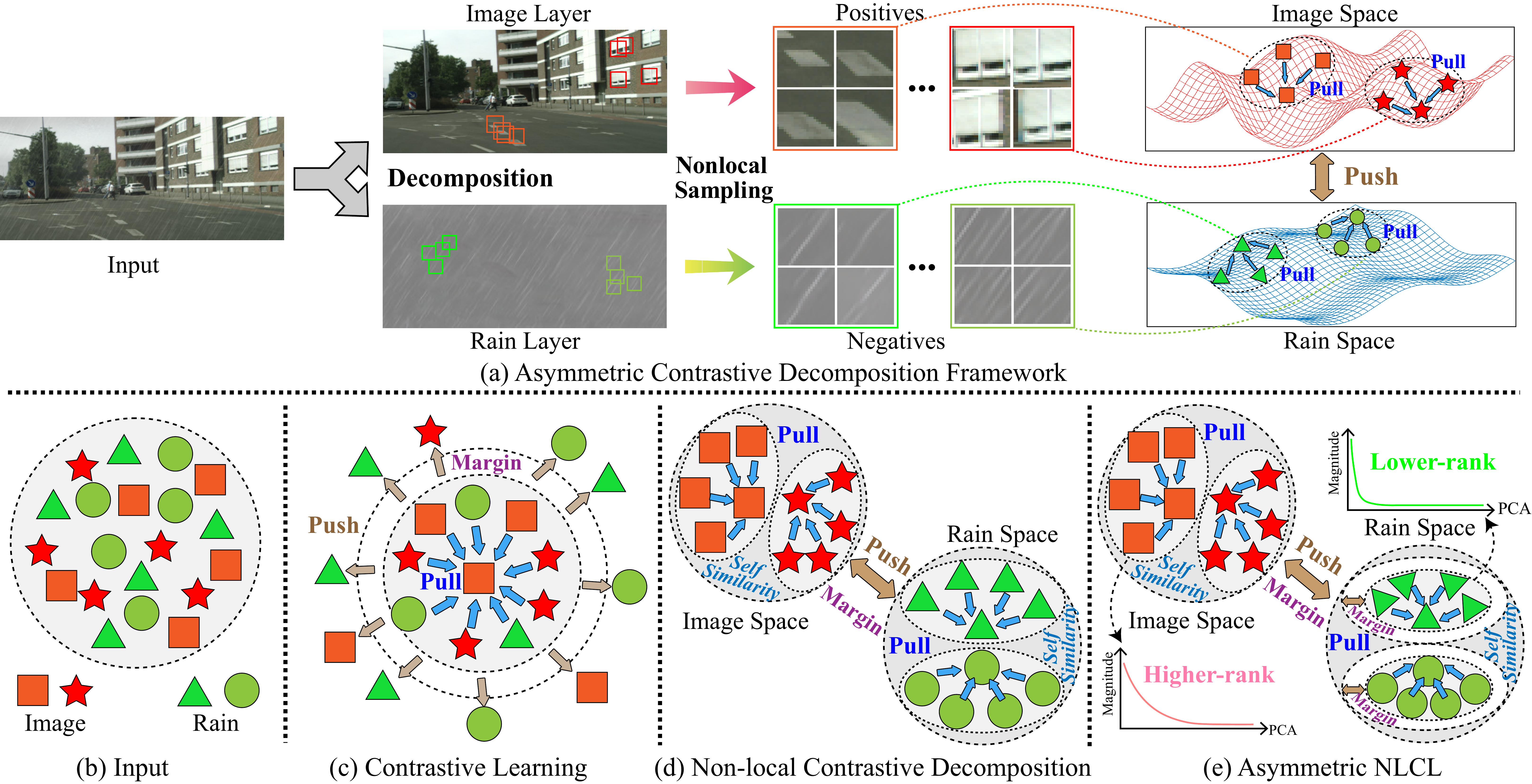}
  \caption{Motivation illustration of the proposed method. (a) Overview of the proposed contrastive decomposition framework for image deraining. Most previous methods model the property of the image layer and rain layer independently in a supervised manner. In this work, we go further by considering the mutually exclusive relationship between the two layers for better decomposition. (b)-(f) illustrate the difference between proposed method and previous contrastive learning. (c) contrastive learning is difficult to well decouple the two components. (d) The NLCL exploits the non-local self-similarity to improve the positive/negative sampling with discriminative representation; on the other hand, the contrastive decomposition would take the bidirectional contrastive learning for better decomposition. (e) In this work, we further take the fine-grained asymmetric property between the rain and image layers into consideration to obtain more compact low-dimensional manifold for each non-local clusters. }
  \label{Motivation}
\end{figure*}

Thus, the image deraining task can be formulated as an ill-posed inverse problem with following optimization function:
\begin{equation}
\setlength{\abovedisplayskip}{2pt}
\setlength{\belowdisplayskip}{2pt}
\mathcal{L}_{decom}  =  ||\textbf{\emph{B}} + \textbf{\emph{R}} - \textbf{\emph{O}}||_F^2 + \delta P_{b}(\textbf{\emph{B}}) + \lambda P_{r}(\textbf{\emph{R}}),
\label{eq:RegularizationFormulation}
\end{equation}
where the first term is self-consistent loss, namely the data fidelity term, $P_{b}$ and $P_{r}$  denote the prior knowledge for the clean image and rain streaks, respectively. Thanks to the sparsity of the rain streaks in space, in this work, we regularize the rain layer with the $L_1$ constraint: $P_{r}(\textbf{\emph{R}}) = ||\textbf{\emph{R}}||_1$ favoring the rain streaks with large discontinuities. On the other hand, for the clean images, we employ the adversarial loss \cite{Goodfellow2014Generative} to learn the distribution mapping differing the rainy image from clean image:
\begin{equation}
\begin{aligned}
\setlength{\abovedisplayskip}{2pt}
\setlength{\belowdisplayskip}{2pt}
P_{b}(\textbf{\emph{B}}) = \mathbb{E}_{\textbf{\emph{B}}}\left [ \textrm{log}D(\textbf{\emph{B}}) \right ] + \mathbb{E}_{\textbf{\emph{O}}}\left [ \textrm{log}(1-D(G_\textbf{\emph{B}}(\textbf{\emph{O}})))\right],
\end{aligned}
\end{equation}
where \emph{D} is the discriminator, and $G_\textbf{\emph{B}}$ is the generator for the clean image. The proposed decomposition-based architecture is shown in Fig. \ref{Motivation}(a), which consists of two branches to restore the background ($G_{\textbf{\emph{B}}}$) and extract the rain ($G_{\textbf{\emph{R}}}$), respectively. Moreover, we enforce the unsupervised loss on the decomposition framework to optimize the decoupled two components. Note that, the proposed method can be regarded as the integration of the model-driven optimization method and data-driven learning network \cite{yu2021unsupervised}with both the good generalization endowed by the unsupervised loss and good representation endowed by the deep network.

Most of the existing restoration methods follow the decomposition framework in Eq. (\ref{eq:RegularizationFormulation}) with different hand-crafted \cite{li2016rain} or learned priors \cite{yu2021unsupervised}, where they only consider the clean image or rain layer separately. That is to say, Equation (\ref{eq:RegularizationFormulation}) mainly focuses on modelling of the statistical property of the signal itself. However, it has neglected the relationship between clean image \textbf{\emph{B}}, rain layers \textbf{\emph{R}}, and observed image \textbf{\emph{O}}. In this work, we argue the contrastive relationship among these components can further help to distinguish them from each other. We introduce the contrastive learning to model the relationship between different components for better decomposition. Thus, the overall objective function including the decomposition constraint and contrastive loss is formulated as:
\begin{equation}
\begin{aligned}
\setlength{\abovedisplayskip}{2pt}
\setlength{\belowdisplayskip}{2pt}
\mathcal{L}_{overall}  =  \mathcal{L}_{decom} +  \mathcal{L}_{contrastive}(\textbf{\emph{B}},\textbf{\emph{R}},\textbf{\emph{O}}).
\label{eq:OverallLoss}
\end{aligned}
\end{equation}

\noindent
\textbf{Motivation of Asymmetric Non-local Contrastive Learning}: The contrastive learning has been preliminarily studied in low-level image dehazing \cite{wu2021contrastive} and deraining \cite{chen2022learning}, as shown in Fig. \ref{Motivation}(c). The existing methods simply take the clean images as the positive and the degraded rainy/hazy images as the negative samples. To differ the clean from the degraded images, these models are enforced to attend on the subtle degradations over the whole image. It is especially difficult for the light rain/haze conditions where the degradations are hard to be observed. Moreover, the degradation is entangled with the clean background in the negative samples. Thus, it is easily for the conventional contrastive learning methods to obtain the over-smooth or leave the residual results as shown in Fig. \ref{Motivation}(c).

In Fig. \ref{Motivation}(d), we show the schematic diagram of our non-local contrastive learning. Compared with previous methods, the proposed method takes the estimated clean image and estimated rain as the positive/negative, and verse vice. The disentangled image and rain would significantly reduce the learning difficulty, since the rain and image layers have distinct yet different features. Moreover, we exploit the intrinsic non-local self-similarity within each layer, such as similar patch sampling strategy would further increase the compactness of the contrastive samples of each layer. Overall, the disentangled sample choice and non-local sampling strategy of the proposed method would naturally enlarge the discrepancy between the inter-class samples and simultaneously enhance the compact intra-class samples for better rain and image decomposition.

\begin{figure*}[t]
  \centering
  \includegraphics[width=1.0\linewidth]{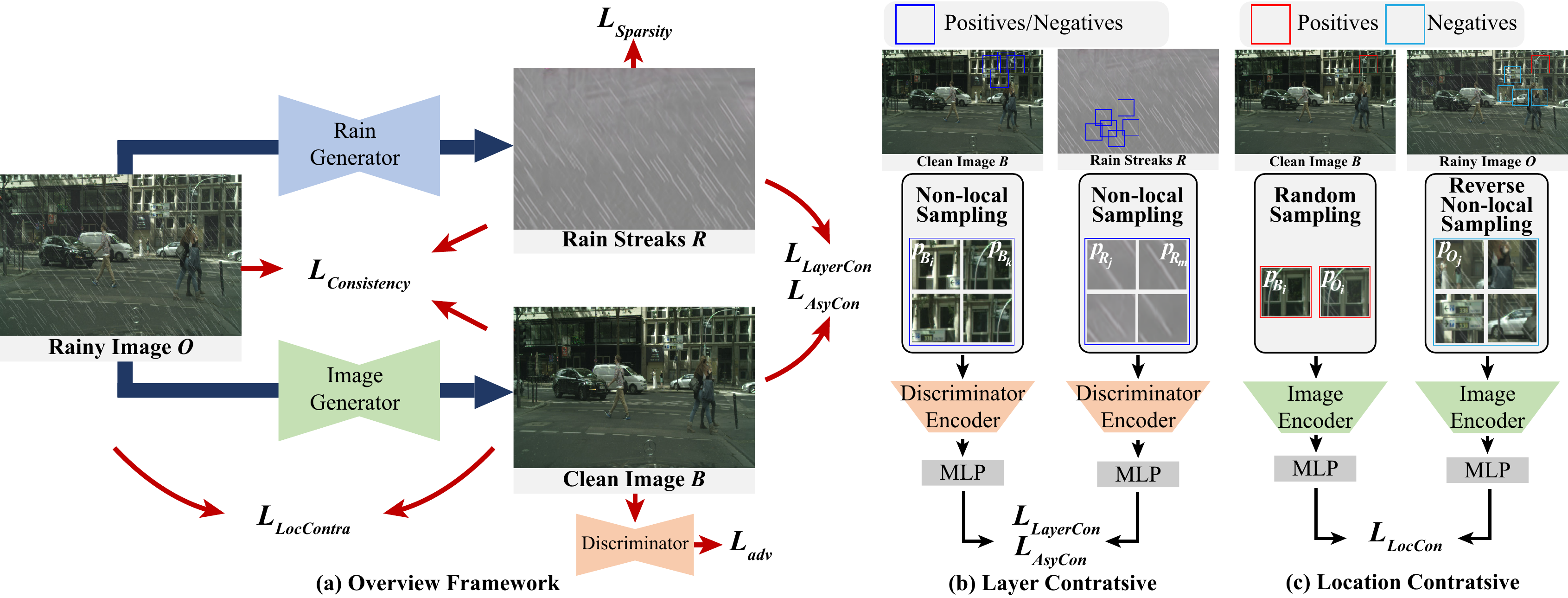}
  \caption{Overall architecture of the proposed method. (a) The ANLCL consists of two sub-networks to extract the background and the rain layers respectively with three contrastive losses. (b) The layer contrastive loss including both the bidirectional (symmetric) layer contrastive and asymmetric layer contrastive between the image and rain layer with the non-local sampling for the decomposition. (c) The location contrastive between the clean image and rainy image with reverse non-local sampling for negatives for image content preserving.}
  \label{Structure}
\end{figure*}

In Fig. \ref{Motivation}(e), we further extend the NLCL to its asymmetric version. The NLCL equally treats the rain and image layer within the non-local contrastive decomposition. However, this is not true, since the image patch contains diverse patterns, such as the complex textures and sharp edges while the rain patches are much more simpler with line-pattern streaks or veiling. That is to say, the image patches space may lie on higher-dimensional manifold while the rain patches space may lie on lower-dimensional manifold with high-probability (Detailed analysis in Section \ref{asymmetric section}). This intuitive observation motivates us to model the compactness difference cue of the rain and image patch into the contrastive decomposition, so as to better differ the rain from image.

The overview architecture of the proposed method is shown in Fig. \ref{Structure}(a). Next, we will first describe how we construct the contrastive relationship among each component \textbf{\emph{B}}, \textbf{\emph{R}}, and \textbf{\emph{O}} (Section \ref{Layer-Location_Contrastive}). Finally, we will presents positive/negative sampling strategy (Section \ref{sampling section}) and design encoders in details (Section \ref{encoders section}).

\subsection{Joint Layer and Location Contrastive}
\label{Layer-Location_Contrastive}
\noindent
\textbf{Layer Contrastive}: First, the clean image \textbf{\emph{B}} and the rain layer \textbf{\emph{R}} are vastly different, in which the rain streaks are simple and directional line-pattern, while the natural images are complex yet meaningful structures such as edges and textures. The \emph{dissimilarity} between the \textbf{\emph{B}} and \textbf{\emph{R}}, as two different categories, can be well modelled by CL as negative pairs. And it is very reasonable to take the patches in the same image as the positive samples. The important sampling strategy and encoder in CL will be discussed in next subsection. In Fig. \ref{Structure}(b), we consider the image and rain layer as the equally symmetric two components. Thus, referring to the rain patches $\textbf{\emph{p}}_{\textbf{\emph{R}}_\textbf{\emph{i}}}$ as negatives, while the background patches as the positives $\textbf{\emph{p}}_{\textbf{\emph{B}}_\textbf{\emph{j}}}$, and vice verse. We propose the bidirectionally symmetric layer contrastive learning between the two layers \textbf{\emph{B}} and \textbf{\emph{R}} which can be formulated as:
\begin{equation}
\begin{aligned}
\setlength{\abovedisplayskip}{2pt}
\setlength{\belowdisplayskip}{2pt}
\mathcal{L}_{LayerCon} = -\frac{1}{N_B}\sum_{k=1}^{N_B}\sum_{i=1}^{N_B}\frac{\textrm{exp}({\textbf{\emph{f}}}_{\textbf{\emph{B}}_\textbf{\emph{i}}}\cdot {\textbf{\emph{f}}}_{\textbf{\emph{B}}_\textbf{\emph{k}}}/\tau)}{\sum_{j=1}^{N_R}\textrm{exp}({\textbf{\emph{f}}}_{\textbf{\emph{B}}_\textbf{\emph{i}}}\cdot {\textbf{\emph{f}}}_{\textbf{\emph{R}}_\textbf{\emph{j}}}/\tau)}\\
-\frac{1}{N_R}\sum_{m=1}^{N_R}\sum_{j=1}^{N_R}\frac{\textrm{exp}({\textbf{\emph{f}}}_{\textbf{\emph{R}}_\textbf{\emph{j}}}\cdot {\textbf{\emph{f}}}_{\textbf{\emph{R}}_\textbf{\emph{m}}}/\tau)}{\sum_{i=1}^{N_B}\textrm{exp}({\textbf{\emph{f}}}_{\textbf{\emph{R}}_\textbf{\emph{j}}}\cdot {\textbf{\emph{f}}}_{\textbf{\emph{B}}_\textbf{\emph{i}}}/\tau)},
\label{eq:LayerCon}
\end{aligned}
\end{equation}
where $\textbf{\emph{f}}_{\textbf{\emph{B}}_\textbf{\emph{i}}} = E_{D}(\textbf{\emph{p}}_{\textbf{\emph{B}}_\textbf{\emph{i}}})$, $\textbf{\emph{f}}_{\textbf{\emph{R}}_\textbf{\emph{j}}} = E_{D}(\textbf{\emph{p}}_{\textbf{\emph{R}}_\textbf{\emph{j}}})$, $\tau$ denotes the scale temperature parameter \cite{chen2020simple}. The first term in Eq. (\ref{eq:LayerCon}) is the image positive and rain negative layer contrastive, and the second term is the rain positive and image negative layer contrastive. Here, we regard the image and rain layer as two equal components. Thus, the bidirectional layer contrastive could facilitate us to better push one layer away from each other, and pull each layer further to different clusters. $E_{D}$ is the encoder of contrastive network. The features $\textbf{\emph{f}}_{\textbf{\emph{B}}_\textbf{\emph{k}}}$ are extracted from the non-local patches $\textbf{\emph{p}}_{\textbf{\emph{B}}_\textbf{\emph{k}}}$ of $\textbf{\emph{p}}_{\textbf{\emph{B}}_\textbf{\emph{i}}}$, while the $\textbf{\emph{f}}_{\textbf{\emph{R}}_\textbf{\emph{m}}}$ are extracted from the non-local patches $\textbf{\emph{p}}_{\textbf{\emph{R}}_\textbf{\emph{m}}}$ of $\textbf{\emph{p}}_{\textbf{\emph{R}}_\textbf{\emph{j}}}$. $N_B$ and $N_R$ denote the sample numbers of positives and negatives.

\noindent
\textbf{Location Contrastive}: Second, we can observe that the clean image \textbf{\emph{B}} and the observed image \textbf{\emph{O}} are visually close to each other, since the rain streaks \textbf{\emph{R}} are much simpler than \textbf{\emph{B}}. The \emph{similarity} between patches of the same location in \textbf{\emph{B}} and \textbf{\emph{O}}, as the same view, can be well modelled as the positive samples. Consequently, we set the patches with different locations as the negative samples. In Fig. \ref{Structure}(c), for location contrastive, there should be only one positive sample, since the location correspondence is exactly one-to-one. The encoder of image generator $E_{G_{\textbf{\emph{B}}}}$ is utilized to extract the patch features, denoted as ${\textbf{\emph{v}}}_{\textbf{\emph{O}}_\textbf{\emph{i}}} = E_{G_{\textbf{\emph{B}}}}(\textbf{\emph{p}}_{\textbf{\emph{O}}_\textbf{\emph{i}}})$, and ${\textbf{\emph{v}}}_{\textbf{\emph{B}}_\textbf{\emph{i}}} = E_{G_{\textbf{\emph{B}}}}(\textbf{\emph{p}}_{\textbf{\emph{B}}_\textbf{\emph{i}}})$. Thus, the location contrastive loss is formulated as:
\begin{equation}
\begin{aligned}
\setlength{\abovedisplayskip}{2pt}
\setlength{\belowdisplayskip}{2pt}
&\resizebox{0.865\hsize}{!}{$\mathcal{L}_{LocCon} = \sum_{i=1}^{N}\frac{\textrm{exp}({\textbf{\emph{v}}}_{\textbf{\emph{O}}_\textbf{\emph{i}}}\cdot {\textbf{\emph{v}}}_{\textbf{\emph{B}}_\textbf{\emph{i}}}/\tau)}{\textrm{exp}({\textbf{\emph{v}}}_{\textbf{\emph{O}}_\textbf{\emph{i}}}\cdot {\textbf{\emph{v}}}_{\textbf{\emph{B}}_\textbf{\emph{i}}}/\tau)+   \sum_{j=1}^{N}\textrm{exp}({\textbf{\emph{v}}}_{\textbf{\emph{O}}_\textbf{\emph{j}}}\cdot {\textbf{\emph{v}}}_{\textbf{\emph{B}}_\textbf{\emph{i}}}/\tau)},$}
\end{aligned}
\end{equation}
where $N$ is the negative sample numbers. The location contrastive constrains the restored background patches $\textbf{\emph{p}}_{\textbf{\emph{B}}_\textbf{\emph{i}}}$ at location $i$ to be related (positive) with the corresponding input patches $\textbf{\emph{p}}_{\textbf{\emph{O}}_\textbf{\emph{i}}}$ in comparison to other random patches $\textbf{\emph{p}}_{\textbf{\emph{O}}_\textbf{\emph{j}}}$, so as to retain the image content. Overall, the layer contrastive is to remove the rain from the image layer, while the location contrastive is to preserve the image content in the removal. The two contrastive losses compete with each other to obtain the balance.

\begin{figure*}[t]
  \centering
  \setlength{\abovecaptionskip}{5pt}
  \setlength{\belowcaptionskip}{-5pt}
   \includegraphics[width=1.0\linewidth]{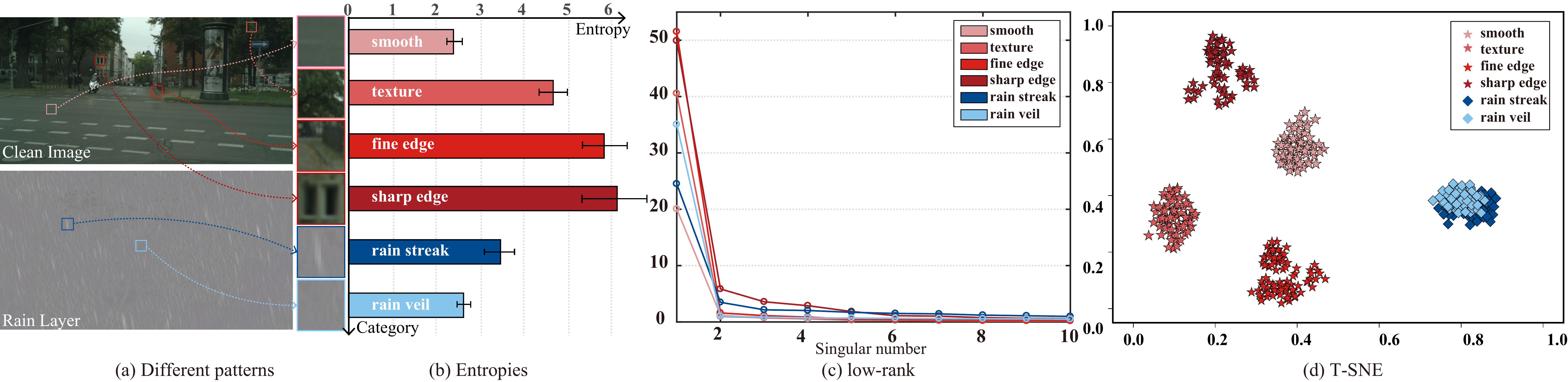}
  \caption{Illustration of the asymmetry between the image and rain layers. (a) Clean image and rain layers. We select the typical patches from both of them, such as the smooth, texture, edges regions of the image, and rain streak and veiling regions of the rain. Each category is with 100 samples. (b) The entropy of each category patches. (c) The singular value decomposition of each category patches. (d) The T-SNE visualization of each category patches. It has been shown that the intrinsic dimension of the rain and image patches share obvious dissimilarity.}
  \label{Asymmetry}
\end{figure*}

\subsection{Asymmetric Layer Contrastive}
\label{asymmetric section}
\noindent
Although the joint layer and location contrastive decomposes image and rain layer by intrinsic feature discrepancy, the compactness discrepancy between each layer is not fully considered. In this work, we extend the symmetric contrastive decomposition to the asymmetric contrastive decomposition. Specifically, we discover that the dimension of image space is generally higher than that of rain space, which reveals that the image patches contain more diverse and complex structures and patterns. In other words, the compactness between rain patches should be tighter than that of image patches. To this end, we employ the entropy, rank and T-SNE as the different ways to both quantitatively and qualitatively analyze both the image and rain patches, which helps to verify the asymmetric property between image and rain layer.

We select the typical 16*16 patches from both clean image layer and rain layer in Fig. \ref{Asymmetry}(a), mainly including six categories: smooth, texture, edges regions of image, and rain streak and veiling regions of the rain. Note that each category contains 100 samples to statistically compute the indexes. We first calculate the entropy of each category patches and make comparison in Fig. \ref{Asymmetry}(b). The entropy is to evaluate the degree of randomness (complexity) in the patch by the definition $-sum(p.*log2(p))$ where \emph{p} is the probability of normalized histogram count. It is obvious that the entropies of edge and texture patches, except for smooth patches, are generally higher than those of rain streak and veiling patches. This is very reasonable since the image layer has diverse and complex structural patterns. We also conduct singular value decomposition (SVD) and T-SNE  visualization of each category patch cluster. In Fig. \ref{Asymmetry}(c), we show the singular value decomposition curve of each category is slightly different from each other. Moreover, the low-rankness of each category is exactly in proportion to the entropy. In Fig. \ref{Asymmetry}(d), we perform the T-SNE to visualize the two-dimensional distribution of these different patches.

This motivates us to utilize the fine-grained discrepancy property between the image and rain which could further offer us abundant cues to better distinguish them. To this end, we extend the symmetric layer contrastive to its asymmetric version by additionally consider the margin loss between the image and rain patches. When sampling image patches with higher entropy than rain patches, we enforce the margin loss between \textbf{\emph{B}} and \textbf{\emph{R}}:
\begin{equation}
\setlength{\abovedisplayskip}{2pt}
\setlength{\belowdisplayskip}{2pt}
\mathcal{L}_{Margin} = {\sum_{i,j=1}^{N_R}\sum_{m,n=1}^{N_B} [\varepsilon + \eta(||{\textbf{\emph{f}}}_{\textbf{\emph{R}}_\textbf{\emph{i}}}- {\textbf{\emph{f}}}_{\textbf{\emph{R}}_\textbf{\emph{j}}})||^2- ||{\textbf{\emph{f}}}_{\textbf{\emph{B}}_\textbf{\emph{m}}}- {\textbf{\emph{f}}}_{\textbf{\emph{B}}_\textbf{\emph{n}}}||^2]_{+}},\\
\label{eq:Margin}
\end{equation}
where $[z]_{+} = max(z, 0)$ denotes the standard hinge loss, $\varepsilon$ is the pre-defined margin usually set as 1. $\eta\in\{1,-1\}$ is to indicate whether the entropy of image patches is larger or smaller than that of the rain patches. The physical meaning of Eq. (\ref{eq:Margin}) is to make the distance of the rain patches maintain a large margin over the distance of the image patches. Note that, in each batch, we need pre-calculate the entropy of the chosen image and rain patches.

In order to accommodate with symmetric contrastive loss, we further translate marginal loss Eq. (\ref{eq:Margin}) into the asymmetry contrastive loss as follow:
\begin{equation}
\begin{aligned}
\setlength{\abovedisplayskip}{2pt}
\setlength{\belowdisplayskip}{2pt}
\mathcal{L}_{AsyCon} = -\frac{1}{N_R}\frac{1}{N_B}\left(\frac{\sum_{i=1}^{N_R}\sum_{j=1}^{N_R}\textrm{exp}({\textbf{\emph{f}}}_{\textbf{\emph{R}}_\textbf{\emph{i}}}\cdot {\textbf{\emph{f}}}_{\textbf{\emph{R}}_\textbf{\emph{j}}}/\tau)}{\sum_{m=1}^{N_B}\sum_{n=1}^{N_B}\textrm{exp}({\textbf{\emph{f}}}_{\textbf{\emph{B}}_\textbf{\emph{m}}}\cdot {\textbf{\emph{f}}}_{\textbf{\emph{B}}_\textbf{\emph{n}}}/\tau)}\right)^\eta,\\
\end{aligned}
\label{eq:AsyCon}
\end{equation}
where the parameters and functions in Eq. (\ref{eq:AsyCon}) have the same meaning as Eq. (\ref{eq:LayerCon}). Note that, the goal of the symmetric contrastive loss in Eq. (\ref{eq:LayerCon}) is to equally minimize the distance within the rain/image space, while enlarge the distance between the rain and image space, while the goal of the asymmetric contrastive loss in Eq. (\ref{eq:AsyCon}) is to regularize the distance between each space according to different contents. The asymmetric contrastive loss would further capture the fine-grained compactness discrepancy and improve the discriminative representation between two layers.

\subsection{Non-local Sampling Strategy}
\label{sampling section}
In contrastive learning, the negatives are the samples which should be discriminated by the learned representations, while the positives are highly related and possess the invariance in the learned representations. The previous methods usually use the augmentations to construct the single instance positives and randomly sampling as the negatives \cite{chen2020simple}. Note that, the self-similarity is a generic and powerful prior knowledge. In this work, we introduce the non-local self-similarity to automatically select both positive and negative samples  within a single image. We employ the block matching \cite{dabov2007image} with $L_2$ Euclidian distance to measure the dissimilarity/similarity in image space:
\begin{equation}\label{eq:PatchMatching}
\setlength{\abovedisplayskip}{2pt}
\setlength{\belowdisplayskip}{2pt}
Dist(\textbf{\emph{p}}_{i}, \textbf{\emph{p}}_{{i}_{R}})  =  || \textbf{\emph{p}}_{i} - \textbf{\emph{p}}_{{i}_{\Omega}}||^2,
\end{equation}
where $\textbf{\emph{p}}_{i}$ is the query patch, $\textbf{\emph{p}}_{{i}_{\Omega}}$ are the searched patches in the support set $\Omega$. We take the top-\emph{k} smallest \emph{Dist}() as the similar patches, while the top-\emph{k} largest \emph{Dist}() can be regarded as the dissimilar patches. On one hand, the non-local sampling with similar structures would greatly ease the learning difficulty. On the other hand, the small perturbation within the similar samples would further improve the diversity. Moreover, the patches cropped from the image itself would provide more reliable representation learning. The non-local sampling strategy can be applied for sampling the positive and negative. Here we briefly describe how we use the non-local sampling in very flexible ways.

\begin{figure*}[t]
  \centering
  \setlength{\abovecaptionskip}{3pt}
  \setlength{\belowcaptionskip}{-3pt}
   \includegraphics[width=1.0\linewidth]{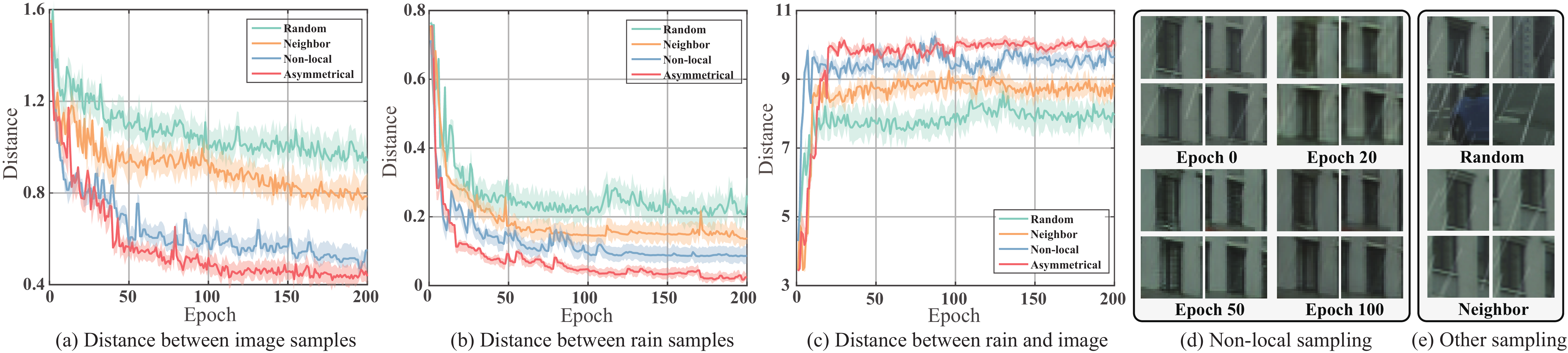}
  \caption{Effectiveness of the non-local sampling strategy. (a) Example of the randomly sampled positives with low similarity in comparison with the non-local self-similar sampled positives. (b) The Euclidean distances of positives decrease rapidly to a relative low level by non-local sampling strategy when compared with the random strategy, indicating the self-similarities are gradually learned and the patches are more relevant in the deraining procedure. (c) The similar patches guide each other to gradually restore the clean image and remove the randomly distributed rains.}
  \label{SamplingStrategy}
\end{figure*}

\noindent
\textbf{Non-local Sampling in Layer Contrastive.} In layer contrastive, the clean image and rain streaks can be regarded as two distinct categories where they have intra-class similarity and inter-class dissimilarity. Our principle is that the positive samples (clean image patches in \textbf{\emph{B}}) should be pulled together as much as possible, so is the negative samples (rain streak patches in \textbf{\emph{R}}) which can also be pulled together. That is to say, we enforce the non-local sampling on both the positive and negative samples. Compared with single positive instance, the multiple non-local positive samples would benefit us to improve the feature representation. The recent research has also shown that positives from multiple instances could improve the representations if sampled appropriately (with supervised labels \cite{khosla2020supervised} or multiple modalities \cite{Han2020VideoRepresentation}). Moreover, compared with the random negative samples, the non-local sampling could additionally model the relationship within the samples.

To illustrate this, Fig. \ref{SamplingStrategy} shows different sampling strategies: random, neighborhood, non-local. The random sampling means we randomly select the patches from the whole image as the positive samples. The neighborhood sampling denotes that the positive samples are sampled from the surrounding patch neighbours with high similarity. Note that, compared with the non-local sampling strategy, the asymmetric non-local is still based-on non-local sampling with additionally asymmetric loss constraint.

In Fig. \ref{SamplingStrategy}(a)-(c), we provide the distance within image patches, distance within rain patches, and distance between the rain and image patches. Compared with random sampling or neighbour sampling, the distances of non-local positives decrease rapidly, and converge at a relatively lower level, which indicates the self-similarities are gradually learned and the patches are more relevant in the restoration procedure. On the contrary, the distance between the rain and image patches are gradually enlarged which means the two components are gradually decoupled. These results convincingly validates superiority of the proposed non-local sampling strategy. Second, the proposed asymmetric loss could further reduce the intra-class similarity and inter-class dissimilarity than that of the non-local sampling. This is because that the asymmetric loss would adaptively learn the compactness distance for each cluster. In addition, we show the progressive deraining results in Fig. \ref{SamplingStrategy}(d) and \ref{SamplingStrategy}(e) for non-local and other sampling. With the increasing epoch, the rain and image are gradually decoupled while the neighborhood sampling would leave slightly residual rain streak in the image.

\noindent
\textbf{Non-local Sampling in Location Contrastive.} The observed image \textbf{\emph{O}} and clean image \textbf{\emph{B}} are very similar to each. In location contrastive, the goal is to retain the image content and remove the rain streaks in observed image, which is exactly a image-to-image translation task.
Thus, we follow the CUT \cite{park2020contrastive} by setting the patches of the same location in \textbf{\emph{B}} and \textbf{\emph{O}} as the positive samples with a large batch size. The previous methods including CUT randomly select the different patches as the negative. However, it is more reasonable that the more dissimilar from the positive sample, the better the negative sample is. This motivates us to still use the non-local sampling strategy to construct the negative patches. Instead of calculating the nearest top-\emph{k} samples, we choose the farthest top-\emph{k} samples (the largest distance) which means they are mostly different from the target positive. We name this negative sampling as the reverse non-local sampling.

\begin{figure}[t]
  \centering
  \includegraphics[width=1.0\linewidth]{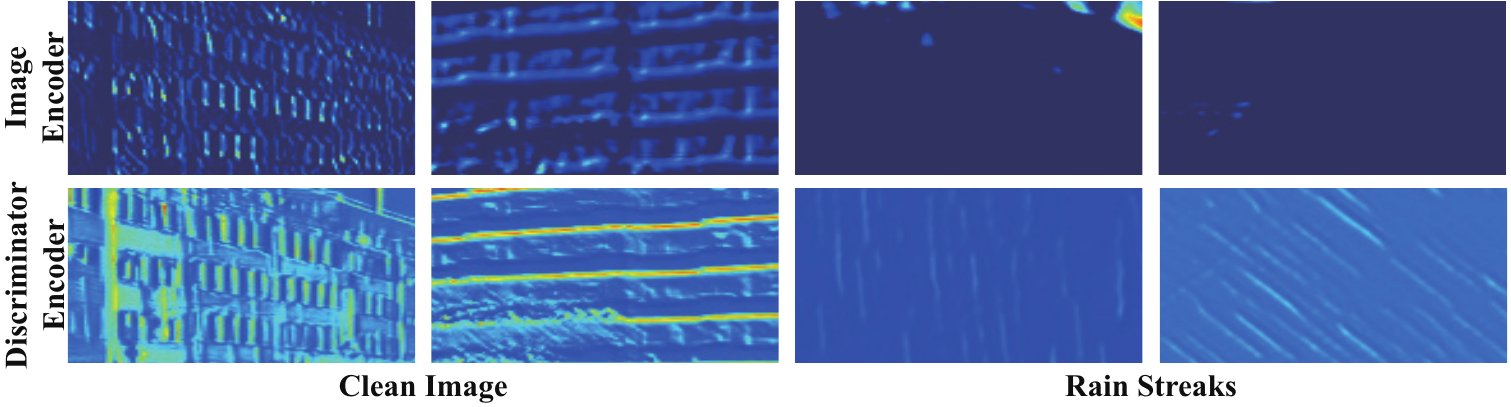}
  \setlength{\abovecaptionskip}{3pt}
  \setlength{\belowcaptionskip}{-3pt}
  \caption{Effectiveness of the discriminator encoder. The first row shows the features extracted from image encoder. Although the extracted features in two different images are clear, the image generator has nearly no response to the rain streaks. The second row shows the features extracted from the discriminator encoder. The extracted features in two images and rain streaks are both clear and discriminative. This strongly supports the effectiveness of the discriminator serving as the encoder for the image and rain layers.}
  \label{FeatureEncoder}
\end{figure}

\begin{figure*}[t]
  \centering
  {\includegraphics[width=1.0\linewidth]{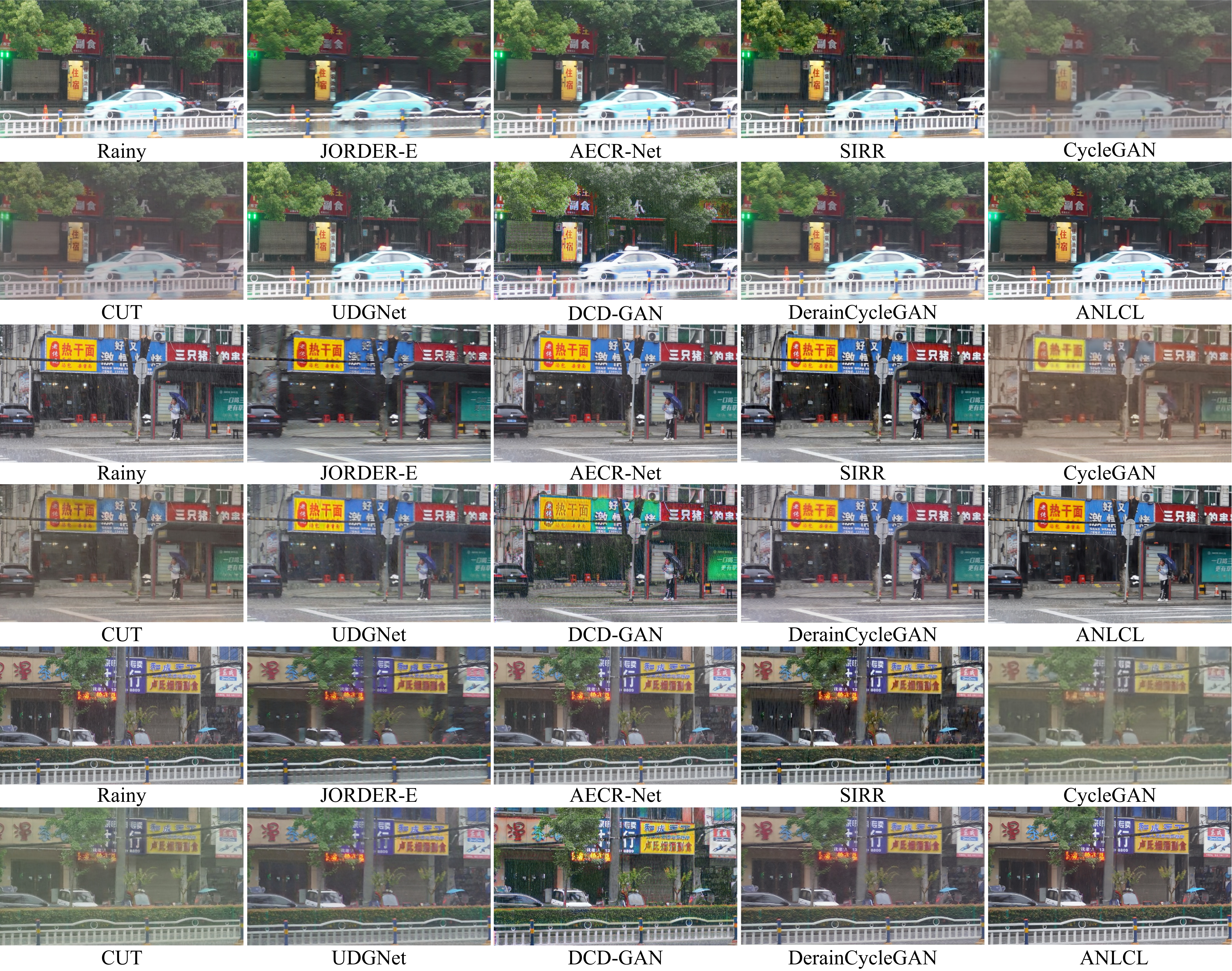}}
	\caption{Visual comparisons in real rainy scenes including both  rain streaks and heavy veiling. We suggest to view the zoomed results on PC.}
	\label{Real}
\end{figure*}

\subsection{Choice of Feature Encoder}
\label{encoders section}
In contrastive learning, the feature encoder is to map the inputs to the embedding low-dimensional feature representation space that facilitates the measurement of the distances between positive and negative samples. It has been recognized that for different CL tasks, the choice of the encoder would vastly influence the final performance \cite{le2020contrastive}. In this work, we also demonstrate that the encoder is indeed tasks dependent for low-level restoration tasks, and explore different encoders for both the layer and location contrastive constraints intuitively and experimentally.

As for the layer contrastive, the goal is to differ the rain streaks from the clean image, which has been analyzed that this is analog to a classification problem. That is to say, the encoder of the layer contrastive should extract the high-level semantic about the category information. The discriminator is in line with the layer contrastive encoder, which can differ the image from non-image component including the rain streaks. As for the location contrastive, the clean image and the observed rainy image are very similar to each other, in which the clean image is the dominant component in rainy image. In other words, the encoder of the location contrastive should well extract the image features. The image generator can satisfactorily achieve this goal.

\begin{table}[]
  \centering
  \footnotesize
  \caption{Quantitative comparisons with SOTA unsupervised methods on synthetic and real datasets.}
  \renewcommand\arraystretch{1.05}
  \setlength{\tabcolsep}{1.8mm}{
   \begin{tabular}{c|c|c|c|c|c|c}
   \hline
   \multirow{2}{*}{Methods} & \multicolumn{3}{c|}{RainCityscapes \cite{hu2019depth}} & \multicolumn{3}{c}{SPA \cite{wang2019spatial}} \\
   \cline{2-7}
   & PSNR & SSIM & NIQE  & PSNR & SSIM & NIQE \\
   \hline
    DSC \cite{luo2015removing}			& 24.91 & 0.7603 & 6.17 & 33.71 & 0.9127 & 9.82 \\
   DIP \cite{ulyanov2018deep}			& 22.45 & 0.6936	& 7.86 & 30.36 & 0.8422 & 9.97\\
   CycleGAN \cite{zhu2017unpaired}	& 24.86 & 0.7906 &3.68 & 33.54 & 0.9127 & \textbf{6.67} \\
   UDGNet \cite{yu2021unsupervised}	& 25.16 & 0.8749 &	5.31  & 29.67 & 0.9299 & 9.50 \\
   CUT \cite{park2020contrastive}	& 25.21 & 0.8225 & 4.08 & 32.97 & 0.9434 & 9.60 \\
   DCD-GAN \cite{chen2022unpaired}	&  25.18 & 0.8270 & 3.73 & 29.23 & 0.9195 & 8.47 \\
   NLCL \cite{ye2022unsupervised}	& 26.46 & 0.8666 & \textbf{3.67} & 33.82 & 0.9468 & 9.55 \\
   DeCycleGAN \cite{wei2019deraincyclegan}	& 26.99 & 0.8670 & 4.86 & 34.16 & 0.9436 & 9.02 \\
   ANLCL 	& \textbf{27.42} & \textbf{0.9123} & 3.72 & \textbf{35.07} & \textbf{0.9505} & 8.46 \\\hline
  \end{tabular}}
  \label{quanCompare}
 \end{table}

To verify our hypothesis, Fig. \ref{FeatureEncoder} visualizes the embedded features map encoded by different encoders: image generator and discriminator. The first row shows the features extracted from image encoder, and the second row shows the features extracted from discriminator encoder. We select two different clean images and two different rain streaks as the example. We can observe that the image generator could effectively extract the image structures, while it cannot extract any informative information from the rain streaks. On the contrary, the line patterned rain streaks and image structure can be clearly observed in the features extracted by the discriminator encoder. The discriminator focuses on the distinguishable features of image and non-image factors to perform the classification task, which matches the layer contrastive learning task better.

\begin{figure*}[t]
  \centering
  \includegraphics[width=1.0\linewidth]{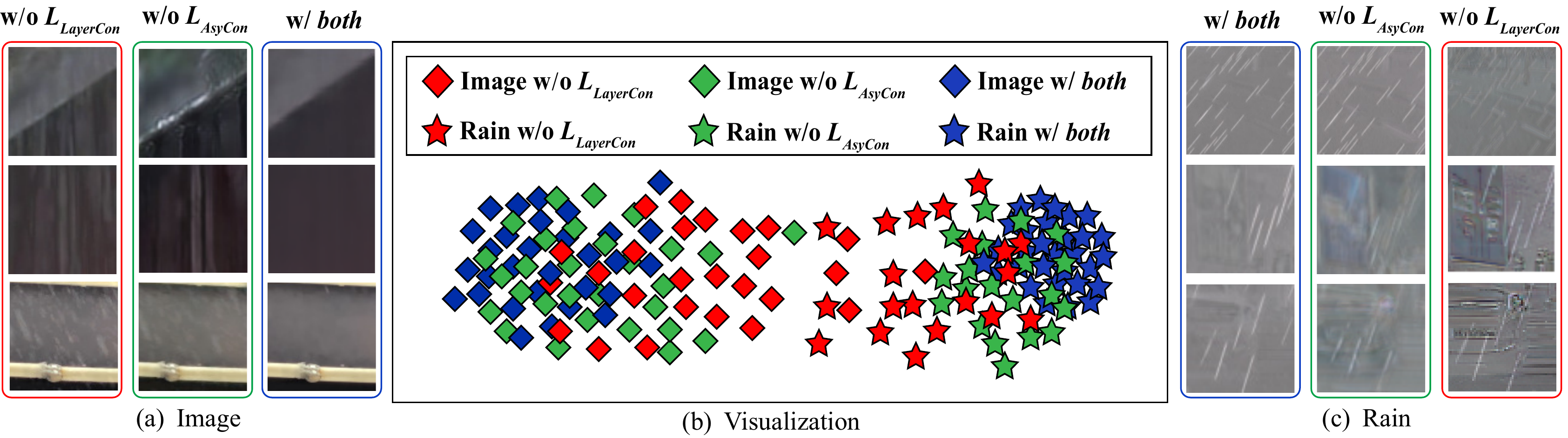}
  \caption{The effectiveness of the layer contrastive for better image and rain streaks decomposition. (a) and (c) show the decoupled rain and image patches w/o the layer contrastive and asymmetric layer contrastive, w/o asymmetric layer contrastive only and w/ both layer contrastive and asymmetric layer contrastive, respectively. (b) visualizes the low-dimensional distributions of image and rain samples in different conditions.}
  \label{Decomposition}
\end{figure*}

\subsection{Implementation Details}
The proposed method facilitates us to unsupervisedly differentiate the rain from clean image. However, training ANLCL model with random initialization may not disentangle the image layer from rain layer. To solve this problem, we employ the two-stage pretrain and finetune training strategy \cite{yu2021unsupervised}. First, we utilize supervised deraining knowledge based on synthetic datasets as the initialization of both generators, and train the generators and discriminator without the CL loss. Second, we train the whole ANLCL model with overall losses using Eq. (\ref{eq:OverallLoss}). Since the CL encoder is associated with the discriminator, the contrastive learning loss will not work until the well-trained discriminator is obtained.

We utilize the same ResNet architectures \cite{johnson2016perceptual} for both the image and rain feature extraction. PatchGAN \cite{isola2017image} is employed as the discriminator. We first calculate the top-\emph{k} non-local patches in image space, then obtain the multilayer features \cite{park2020contrastive} from the encoder, and finally embed the non-local features through a two-layer MLP with 256 units. The sampling number $N, N_B$, and $N_R$ are set as 256, 8, 256. The encoder updating follows the setting of MoCo \cite{he2020momentum}, using momentum value $0.99$ and temperature $0.77$. The balance weights for each loss $\lambda$, $\delta$, $\mu$, $\sigma$ and $\gamma$ are set as $0.1, 1, 1, 1, 0.01$. Due to the high-resolution of FCRealRain, we first downsample the original image by a factor of 4 and then randomly crop $256 \times 256$ patches for training. We adopt Adam optimizer and train the network with learning rate $0.0001$, and batch size $4$ on four RTX 3090 GPUs.

 \begin{table}[t]
 \small
  \centering
\caption{Ablation on different sampling strategies.}
  \begin{threeparttable}
  \renewcommand\arraystretch{1.0}
  \setlength{\tabcolsep}{1.6mm}{
  \begin{tabular}{cc|cc}
   \hline
   	Positive & Negative & PSNR & SSIM  \\
	\hline	
	Random	&	Random	& 25.83	&	0.8471	\\
	Neighbour	&	Random	&	26.03	&	0.8491	\\
	Neighbour	&	Neighbour	&	25.97	&	0.8477	\\
	Non-local	&	Random	& 26.18	&	0.8489	\\
	Random	& Non-local	&	26.16	&	0.8531	\\
	Non-local	&	Non-local	&	\textbf{26.46}	&	\textbf{0.8666}	\\\hline
  \end{tabular}}
  \end{threeparttable}
  \label{ablationSamplingStrategy}
  \end{table}

\begin{table}[t]
 \small
\centering
  \caption{The choice of different feature encoders.}
  \begin{threeparttable}
  \renewcommand\arraystretch{1.05}
  \setlength{\tabcolsep}{0.5mm}{
  \begin{tabular}{c|ccc}
  \hline
  Encoder & PSNR  & SSIM  & NIQE  \\
  \hline
  Image Generator & 24.86 & 0.8046 & 3.83\\
  Image-Rain Generator & 24.12 & 0.8023 & 3.95 \\
  Discriminator & \textbf{26.46}& \textbf{0.8666} & \textbf{3.67} \\\hline
  \end{tabular}}
  \end{threeparttable}
  \label{ablationFeatureEncoder}
 \end{table}

\section{Experimental Results}

\subsection{Datasets and Experimental Settings}
We conduct the experiments on both synthetic dataset RainCityscapes \cite{hu2019depth}, real dataset SPA \cite{wang2019spatial} and proposed FCRealRain. To simulated the real situation, we split the RainCityscapes with 1400 for training and 175 for testing. Note that we have no access to the ground truth and can only learn in an unsupervised manner. For the real SPA dataset, we obtain 2000 rainy images from SPA for training and 200 rainy images for testing. For a fair comparison, we mainly select the unsupervised methods, including the optimization-based DSC \cite{luo2015removing}, CNN-based DIP \cite{ulyanov2018deep}, GAN-based CycleGAN \cite{zhu2017unpaired} and DerainCycleGAN \cite{wei2021deraincyclegan}, contrastive learning-based CUT \cite{park2020contrastive} and DCD-GAN \cite{chen2022learning}, and optimization-driven deep CNN \cite{yu2021unsupervised}. Furthermore, we compare with state-of-the-art supervised JORDER-E \cite{yang2020joint} and AECR-Net \cite{wu2021contrastive} on the real rainy images. We employ full-reference PSNR/SSIM and no-reference natural image quality evaluator (NIQE) \cite{mittal2012making} to evaluate the deraining performance. Moreover, we employ the mean Average Precision (mAP) to evaluate the downstream object detection for comprehensive evaluation.

 \begin{table}[t]
 \small
 \centering
  \caption{Effectiveness of each loss in ANLCL.}
  \begin{threeparttable}
  \renewcommand\arraystretch{1.0}
  \setlength{\tabcolsep}{0.9mm}{
  \begin{tabular}{cccccc|cc}
  \hline
  $\mathcal{L}_{1}$ & $\mathcal{L}_{adv}$ & $\mathcal{L}_{MSE}$ & $\mathcal{L}_{LocCon}$&$\mathcal{L}_{LayerCon}$&$\mathcal{L}_{AsyCon}$&PSNR&SSIM \\
  \hline
  - &  \checkmark & \checkmark &  \checkmark & \checkmark & \checkmark &27.07&0.9078 \\
   \checkmark & - &  \checkmark &  \checkmark & \checkmark & \checkmark &22.79&0.7948 \\
   \checkmark &  \checkmark & - &  \checkmark & \checkmark & \checkmark &26.91&0.8930 \\
  \checkmark & \checkmark & \checkmark & - &\checkmark &\checkmark &26.45&0.8797 \\
 \checkmark & \checkmark & \checkmark & \checkmark & - &\checkmark &25.83&0.8733 \\
  \checkmark & \checkmark & \checkmark &\checkmark &\checkmark &- &26.46&0.8666 \\
 \checkmark & \checkmark & \checkmark &\checkmark & - & - &24.98&0.8426 \\
 \checkmark & \checkmark & \checkmark & - & - & - &23.55&0.8132 \\
  \checkmark & \checkmark & \checkmark &\checkmark &\checkmark &\checkmark &\textbf{27.42}&\textbf{0.9123} \\
\hline
  \end{tabular}}
  \end{threeparttable}
  \label{ablationLosses}
 \end{table}

\subsection{Comparisons with State-of-the-arts}
In Table \ref{quanCompare}, we report the quantitative results on RainCityscape and SPA, respectively. These datasets mainly contains the rain streaks with different visual appearances without the veiling in heavy rainy images. The quantitative results of NLCL mostly outperform competing methods, which verifies the effectiveness of the proposed method. Moreover, the ANLCL has further improved the results and achieved state-of-the-art performance. We emphasize that ANLCL is not designed for the quantitative index on rain streaks. Instead, our philosophy is to unsupervisedly handle the real rains. To validate this, in Fig. \ref{Real}, we compare with the state-of-the-art methods on real-world rainy images of the FCRealRain, which contains both the rain streak and veiling. The unsupervised methods are trained and tested on FCRealRain, while the supervised methods JORER-E and AECR-Net are trained on paired synthetic datasets. The proposed ANLCL consistently achieves more visual pleasing results, which not only remove the rain streaks but also the veiling artifacts meanwhile better preserve the image structure.

\begin{figure*}
  \centering
  {\includegraphics[width=1.0\linewidth]{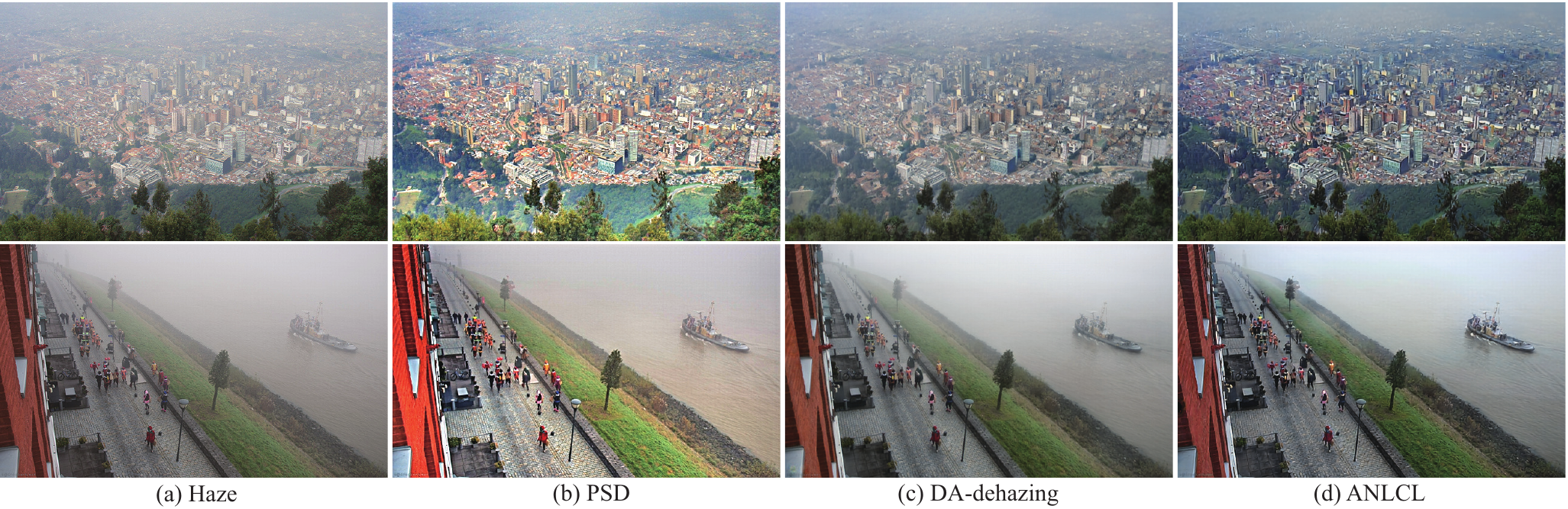}}
	\caption{Visualization comparisons in real haze scenes. ANLCL generalizes well for the hazy image with enhanced contrast and natural appearance.}
	\label{Haze}
\end{figure*}

\begin{figure*}
  \centering
  {\includegraphics[width=1.0\linewidth]{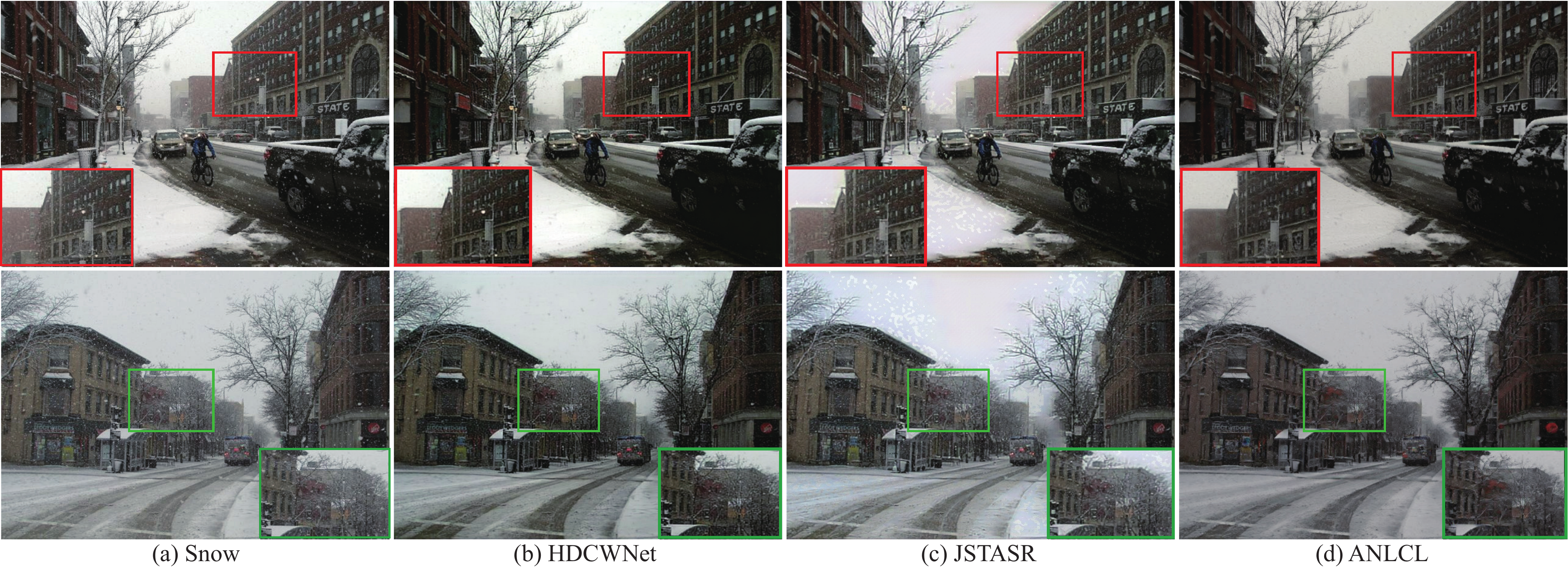}}
	\caption{Visualization comparisons in real snow scenes. ANLCL generalizes well for the snow image with reduced white snow spots.}
	\label{Snow}
\end{figure*}

\subsection{Ablation Study}
\noindent
\textbf{Effectiveness of Non-local Sampling Strategy.} In Table \ref{ablationSamplingStrategy}, we compare the different sampling strategies for both the positives and negatives in contrastive learning, including the random sampling, neighbour sampling (8 nearest neighbour patches), and the proposed non-local sampling. These experiments are all performed on the layer contrastive. Compared with the random sampling, the non-local sampling for both the positive and negative could obviously improve the restoration results. That is to say, the non-local sampling is favorable to learn the image and rain streaks similarity, thus indeed reduces the variance within the positives and negatives, and at the same time enlarge the discrepancy between them. The neighbour sampling could slightly improve the results, while the non-local sampling still obtains the best performance.

\begin{figure}
  \centering
  {\includegraphics[width=1.0\linewidth]{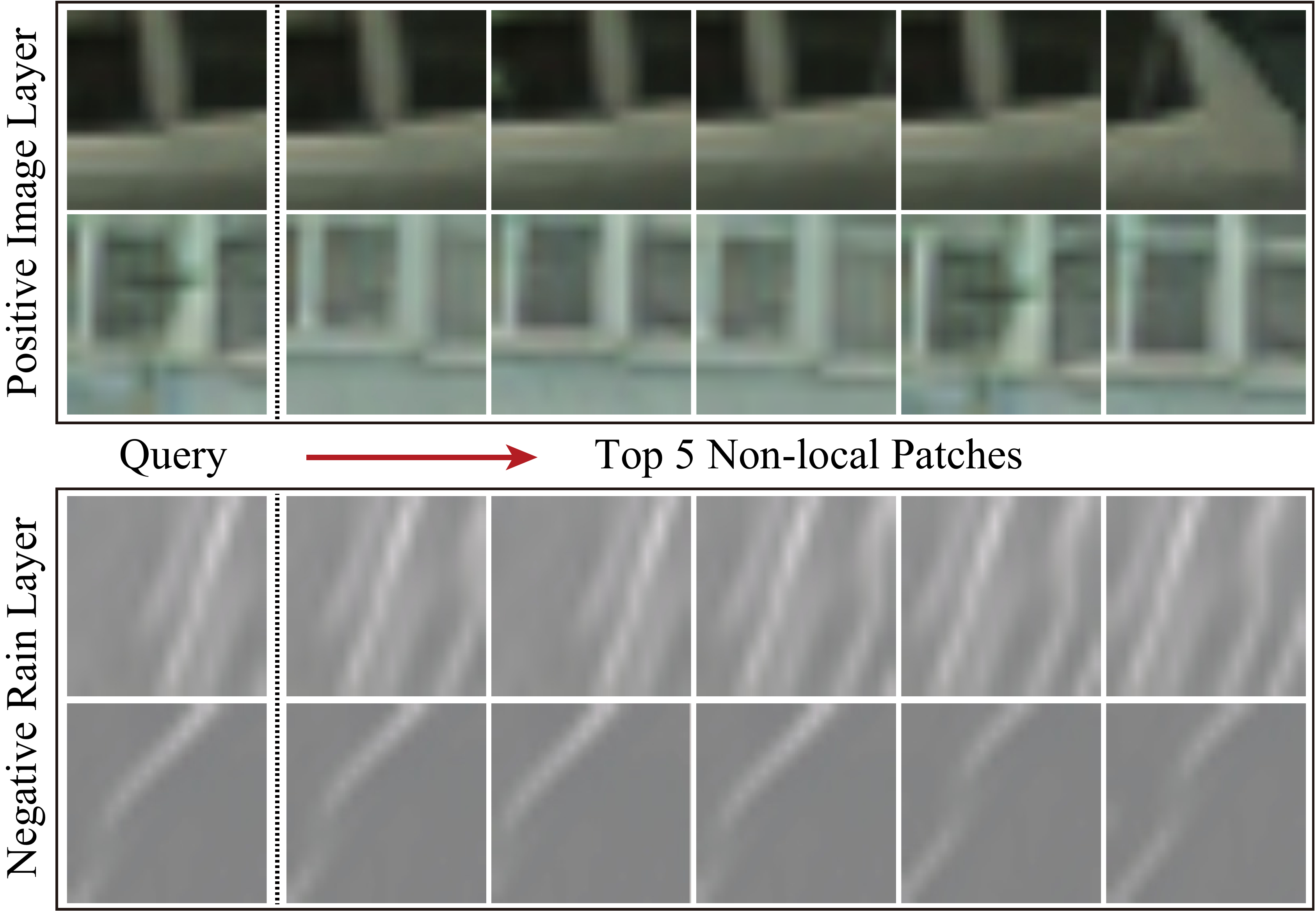}}
	\caption{The visualization of the Top5 non-local searched patches.}
	\label{Non-local}
\end{figure}

\noindent
\textbf{Choice of Different Feature Encoders.} The choice of the encoder for latent feature space is very important. In Table \ref{ablationFeatureEncoder}, we test different encoders for layer contrastive feature embedding. First, we take the image generator as feature encoder for both the image and rain layers. Second, we utilize the image generator and rain generator as feature encoder for the image and rain layer, respectively. Third, we employ the discriminator as the feature encoder for both the image and rain layers. The discriminator encoder has achieved the best result, which verifies the discriminator is suitable to distinguish the image from rain patches. This is reasonable, since the discriminator is trained to differ the real image from the generated image including the artifacts.

\begin{figure*}
  \centering
  {\includegraphics[width=1.0\linewidth]{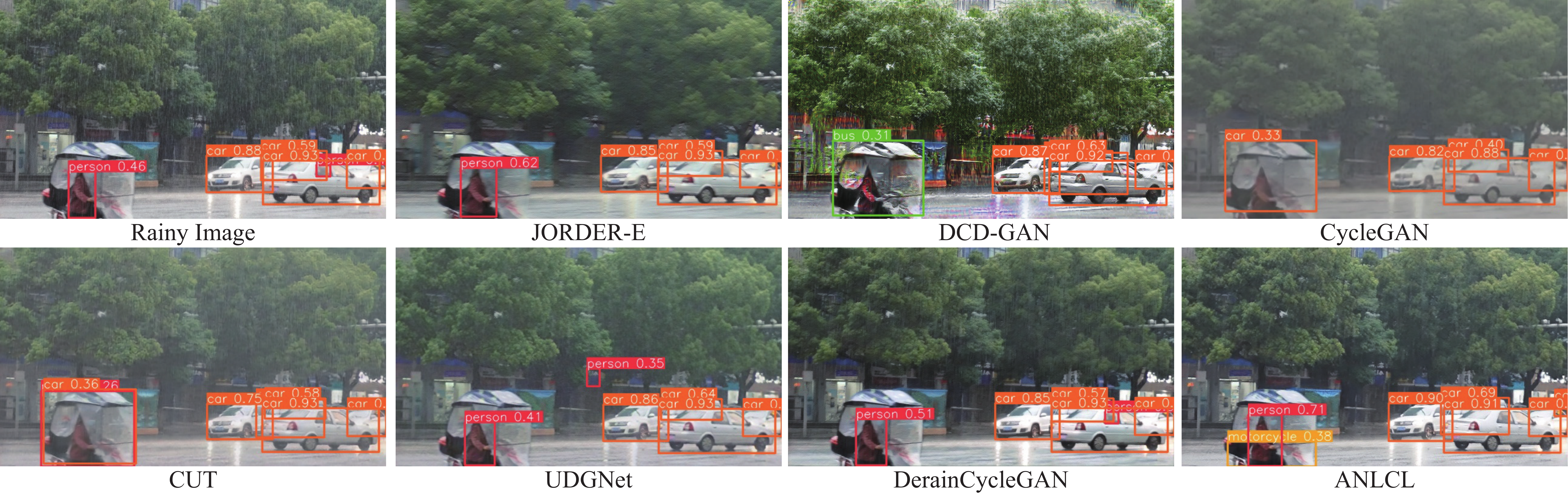}}
	\caption{Visualization comparisons of object detection results after de-raining. Compared with other methods, ANLCL can preserve more background details and boost the performance of detection with higher confidence.}
	\label{detection}
\end{figure*}

\noindent
\textbf{Effectiveness of Each Loss.} In Table \ref{ablationLosses}, we show how each loss contributes to the final result. The $\mathcal{L}_{LocCon}$ and $\mathcal{L}_{LayerCon}$ aim to learn the correlations between the rainy-clean images, and the rain-image layers. We can observe that the two contrastive losses could greatly improve the deraining results, and the self-consistency and adversarial loss are the baseline of our model. $\mathcal{L}_{1}$ sparse loss could slightly improve the performance. The asymmetric contrastive loss $\mathcal{L}_{AsyCon}$ has further improved the results.

\subsection{Analysis and Discussion}
\noindent
\textbf{Effectiveness of Layer Contrastive.} In Fig. \ref{Decomposition}(b), we perform the T-SNE to visualize the distribution of the decomposed image and rain layer w/ and w/o contrastive constraint. Without the layer contrastive, the distribution of the red rhombus (image) and the red pentacle (rain) are divergent. Moreover, they are mixed with each other which means they are still indistinguishable. On the contrary, with the layer contrastive, the distribution of the green rhombus (image) and the green pentacle (rain) are focused and distinguishable. Based on layer contrastive, the proposed asymmetric contrastive further improves the  discriminative representation. Specifically, the distribution of the blue rhombus (image) and the blue pentacle (rain) is well disentangled, and the compactness of blue pentacle (rain) is obviously enhanced comparing with the blue rhombus (image). Moreover, in Fig. \ref{Decomposition}(a) and (c), we visualize several typical decomposition results of both the image and rain patches. It is observed that the layer contrastive can gradually facilitate the disentanglement between the rain and image layers.

  \begin{table}[t]
  \footnotesize
 \centering
  \caption{Boosting performance on supervised deraining methods.}
  \begin{threeparttable}
  \renewcommand\arraystretch{1.0}
  \begin{tabular}{c|ccc}
  \hline
  Method & Baseline & Baseline+ANLCL & Gain \\
  \hline
  PReNet \cite{ren2019progressive}				& 35.16 / 0.9762 & 36.34 / 0.9814	& +1.18 / +0.52\% \\
  RCDNet	\cite{Wang2020Model}			& 36.65 / 0.9805 & 37.46 / 0.9869	& +0.81 / +0.64\% \\
 \hline
  \end{tabular}
  \end{threeparttable}
  \label{boosting}
 \end{table}

 \begin{table}[t]
  \footnotesize
 \centering
  \caption{Boosting performance of object detection task on FCRealRain dataset.}
  \begin{threeparttable}
  \renewcommand\arraystretch{1.0}
  \begin{tabular}{C{1.50cm}|C{0.6cm}|C{0.6cm}C{0.5cm}C{0.50cm}C{0.6cm}C{0.60cm}C{0.60cm}}
    \hline
  \textbf{Method} & mAP & person & car  & motor & bus & \begin{tabular}[c]{@{}c@{}}traffic\\light\end{tabular} & \begin{tabular}[c]{@{}c@{}}traffic\\sign\end{tabular}\\
  \hline
 Rainy Image		& 67.6 & \textbf{90.1} & 93.7 & 46.6	 & 99.4 & 36.7	 & 39.1\\
 JORDER-E		& 63.9 & 85.6 & 93.4 & 39.2	 & 99.5 & 31.8 	 & 33.9\\
 CycleGAN			& 58.8 & 73.8 & 88.9 & \textbf{48.7}	 & 82.8 & 30.7	 & 27.9\\
 CUT					& 61.4 & 85.6 & 93.2 & 27.7	 & 99.1 & 32.5	 & 30.3\\
 UDGNet				& 64.3 & 85.9 & 92.5 & 40.1	 & 99.3 & 33.8	 & 34.2\\
 DCD-GAN       &67.1    &  82.4   & 92.3 &  46.3  &  99.5  &42.2 & 39.9\\
 DeCycleGAN		& 68.7 & 89.5 & 94.0 & 48.1	 & 99.5 & 41.4	 & 39.7\\
 NLCL		& 69.1 & 89.6 & 94.8 & 48.6	 & 99.6 & 45.0	 & 40.1\\
 ANLCL				& \textbf{69.8} & 89.9 & \textbf{95.0} & 48.2	 & \textbf{99.7} & \textbf{45.5}	 & \textbf{40.5}\\
 \hline
  \end{tabular}
  \end{threeparttable}
  \label{boosting on detection}
 \end{table}

\noindent
\textbf{Generalization to Haze and Snow.} The ANLCL is a general prior for image decomposition, in which we do not rely on the specific domain knowledge but exploit the intrinsic self-similarity within each layer and also the discrepancy across different layers. Here, we demonstrate that the ANLCL can be well applied on other typical low-level restoration tasks: haze removal and snow removal. We unsupervisedly re-train the proposed method with real data collected from the Internet, and choose the state-of-the-arts for fair comparison: PSD \cite{chen2021psd} and DA-Dehazing\cite{shao2020domain} for dehazing, HDCWNet \cite{chen2021all} and JSTASR \cite{chen2020jstasr} for desnowing. Note that, both the training and test images are all real without ground truth. In Fig. \ref{Haze}, the ANLCL has obviously enhanced the contrast with natural appearance. In Fig. \ref{Snow}, we remove noticeable snow in the real scenes with the details well preserved.

\noindent
\textbf{Visualization of Self-similarity Patches.} The non-local self-similarity plays the key important role in the unsupervised contrastive learning. We visualize the top 5 non-local positives and negatives of both light and heavy rain conditions in Fig. \ref{Non-local}. It can be observed that the positives and negatives are very similar to that of the query key patch. Compared with other sampling, these self-similarity patches are naturally compact which would significantly facilitates us to learn more discrimination representation. Moreover, it is worth noting that these self-similar patches are real and reliable with slight difference, not the conventional augmentation with fixed transformations. These similar yet different patches are naturally idea samples for the contrastive learning.

\begin{figure}
  \centering
  {\includegraphics[width=1.0\linewidth]{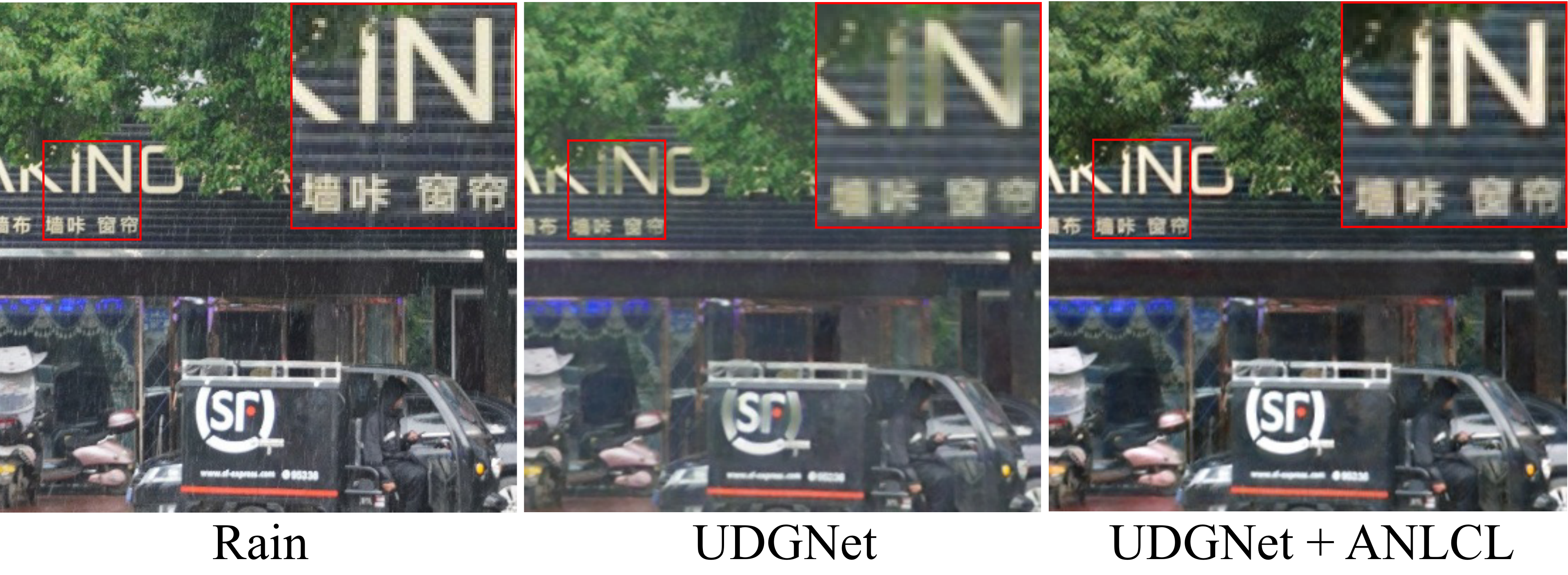}}
	\caption{The benefits of the NLCL when applied to UDGNet.}
	\label{ImprovedUDG}
\end{figure}

\begin{table}[t]
\tiny
 \centering
 \renewcommand\arraystretch{1.0}
 \caption{The model size and inference time under image $256*256$.}
\begin{tabular}{c|c|c|c|c|c|c|c}
\hline
Method   & DSC&JORDER-E & DeCycleGAN & UDGNet & DCD-GAN & CUT & ANLCL   \\ \hline
Size(MB) & -- &16.7     & 43.6     & 5.7   & 11.4 &  45.6   & 2.6    \\ \hline
Time(s)  &33.95 &0.1280    & 0.0379   & 0.0170& 0.0452 &   0.0135  & 0.0098 \\ \hline
\end{tabular}
\label{Time}
\end{table}

\noindent
\textbf{Boosting to Existing Methods.} The ANLCL is a general prior which can be naturally embedded into the existing methods including both the supervised and unsupervised methods. Note that, we only need to enforce the layer contrastive and asymmetric contrastive loss on the existing methods. Thus, the embedded ANLCL loss is easy to be embedded and inference friendly which does not increase any parameters.

For supervised deraining, we choose the state-of-the-art PReNet \cite{ren2019progressive} and RCDNet \cite{Wang2020Model} as example. In Table \ref{boosting}, we list the  PSNR/SSIM performance of Baseline, Baseline+ANLCL and Gain. The existing supervised deraining methods obtain a further improvement after embedded with ANLCL loss. For unsupervised deraining, we take the unsupervised deraining method UDGNet \cite{yu2021unsupervised} as example. In Fig. \ref{ImprovedUDG}, without the ANLCL loss, although UDGNet could well remove the rain streaks, the image structures have been unexpectedly removed along with rain. The result of UDGNet + ANLCL is much better especially for the structure preserving such as the text, which further supports the effectiveness of ANLCL for discriminative image and rain decomposition.

\noindent
\textbf{Promotion for Downstream Detection.} The proposed dataset contains the bounding box annotation for rainy image object detection evaluation. It has been analyzed that not all image deraining methods are beneficial for the detection and tracking \cite{bahnsen2018rain}. In Table \ref{boosting on detection}, we report the mean average precision (mAP) of mentioned results using YOLOv5 on FCRealRain dataset. It is worth noting that JORDER-E, CycleGAN, CUT, UDGNet would lead to negative contribution to the detection results. This is reasonable since the image structure damage outweighes the rain removal benefit in the deraining procedure. On the contrary, the ANLCL consistently improves the detection performance for all categories, except the person. For example, the JORDER-E has unexpectedly over-smoothed the image details which obviously weakens the discriminative feature for detection. In Fig. \ref{detection}, after the rain removal, ANLCL can further detect the motorcycle while other competing methods fails. Moreover, the confidence score has been consistently improved after the ANLCL deraining.

\noindent
\textbf{Model Size and Running Time.} In inference, the proposed method ANLCL contains very simple 9 ResNet blocks, in which we do not employ very complicated architecture. Our goal is to validate the effectiveness of the non-local contrastive loss for discriminative image and rain layer decomposition. In Table \ref{Time}, we report the model size and running time of the competing methods. We can observe that the model size of ANLCL is 2.6M, greatly smaller the other models. Moreover, the running time is 0.01s for ANLCL much faster than other methods, making it more applicable.

\noindent
\textbf{Influence of the Non-local Sampling Number.} We show how the sampling numbers affect the derain result in Table \ref{ablationSamplingSize}. The PSNR increases when the positive sizes grow to an appropriate number, and then decrease since the excessive positives are somehow dissimilar. 8 positives and 256 negatives obtain the best performance. The reason is that most of the rain have the similar line patterns, thus more non-local similar patches can be found to boost the learning than complex image patches. Moreover, the sampling number is not the larger the better, since enforcing the dissimilar patches to be similar may violate the similar assumption.

 \begin{table}[t]
 \small
  \centering
  \caption{The analysis of optimal sampling number.}
  \begin{threeparttable}
  \renewcommand\arraystretch{1.0}
  \setlength{\tabcolsep}{1.5mm}{
  \begin{tabular}{c|cccc}
  \hline
 \diagbox{Pos}{Neg} & 64 & 128 & 256 & 512 \\
 \hline
 4 		& 23.21 & 26.11 & 26.19 & 26.30\\
 8 		& 24.55 & 26.30 & \textbf{26.46} & 26.42 \\
 16 	& 24.54 & 25.96 & 26.02 & 26.11 \\
 32 	& 23.49 & 25.02 & 25.40 & 25.37\\\hline
  \end{tabular}}
  \end{threeparttable}
  \label{ablationSamplingSize}
 \end{table}

\section{Conclusion}
In this paper, we propose a novel asymmetric non-local contrastive learning method for image real rainy reamoval, which explores the powerful self-similarity property within the image. Our unsupervised method can automatically decouple the image from the rain artifacts with good generalization to different real scenes and tasks. We show that our non-local sampling strategy can be used to learn meaningful representations for both positives and negatives. Especially, the proposed non-local sampling strategy enriches the faithful, diverse and structural representation for both negatives and positives. Moreover, we propose asymmetric layer contrastive loss which precisely model the compactness discrepancy for better discriminative decomposition. In addition, we have contributed a high-quality field collection real rain dataset for unsupervised training and testing. Extensive experiments demonstrate that ANLCL achieves state-of-the-art performance in real rainy scenes.

{
\bibliographystyle{ieee_fullname}
\bibliography{egbib}
}

\ifCLASSOPTIONcaptionsoff
  \newpage
\fi

\end{document}